\documentclass[Journal,twocolumn]{IEEEtran}
\usepackage{amsmath}
\usepackage{theorem}
\usepackage{mathrsfs}
\usepackage{accents}
\usepackage{url}
\usepackage{tabularx}
\usepackage{float}
\usepackage{algorithm}
\usepackage{algorithmic}
\usepackage{amsfonts}
\usepackage{amssymb}
\usepackage{mathrsfs}
\usepackage{euscript}
\usepackage{multirow}
\usepackage{cite}
\usepackage{placeins}
\usepackage{color,soul}
\usepackage{soul}
\usepackage{graphicx}
\usepackage{lipsum} 
\usepackage{filecontents} 
\usepackage[caption=true, font=footnotesize]{subfig}
\usepackage{blindtext}
\usepackage{rotating}
\usepackage{balance}
\usepackage{xcolor}

\newcommand{\guanxiong}[1]{\textcolor{black}{#1}}
\newcommand{\hang}[1]{\textcolor{black}{#1}}
\newcommand{\Lee}[1]{\textcolor{black}{#1}}

\begin{document}

\title{Smart Traffic Monitoring System using Computer Vision and Edge Computing}
\author{Guanxiong Liu~\IEEEmembership{Student Member,~IEEE}, Hang Shi, Abbas Kiani~\IEEEmembership{Student Member,~IEEE}, Abdallah Khreishah~\IEEEmembership{Senior Member,~IEEE}, Jo Young Lee, Nirwan Ansari~\IEEEmembership{Fellow,~IEEE}, Chengjun Liu, and Mustafa Yousef

\thanks{Guanxiong Liu, Mustafa Yousef, Abdallah Khreishah and Nirwan Ansari are with the Department of Electrical \& Computer Engineering at New Jersey Institute of Technology, emails: \{gl236,mmy8,abdallah,nirwan.ansari\}@njit.edu, Hang Shi and Chengjun Liu are with the Department of Computer Science at New Jersey Institute of Technology, emails:hs328@njit.edu,cliu@njit.edu, Abbas Kiani is with AT\&T Labs, email: abbas50kiani@gmail.com, and Jo Young Lee is with the Department of Civil Engineering at New Jersey Institute of Technology email: jo.y.lee@njit.edu   
		}
\thanks{Part of this work was presented in the 5th international conference on Internet of Things: Systems, Management and Security, 2018 \cite{kiani2018two}.}
        }
\maketitle
\begin{abstract}
Traffic management systems capture tremendous video data and leverage advances in video processing to detect and monitor traffic incidents. The collected data are traditionally forwarded to the traffic management center (TMC) for in-depth analysis and may thus exacerbate the network paths to the TMC. To alleviate such bottlenecks, we propose to utilize edge computing by equipping edge nodes that are close to cameras with computing resources (e.g. cloudlets). A cloudlet, with limited computing resources as compared to TMC, provides limited video processing capabilities. In this paper, we focus on two common traffic monitoring tasks, congestion detection, and speed detection, and propose a two-tier edge computing based model that takes into account of both the limited computing capability in cloudlets and the unstable network condition to the TMC. Our solution utilizes two algorithms for each task, one implemented at the edge and the other one at the TMC, which are designed with the consideration of different computing resources. While the TMC provides strong computation power, the video quality it receives depends on the underlying network conditions. On the other hand, the edge processes very high-quality video but with limited computing resources. Our model captures this trade-off. We evaluate the performance of the proposed two-tier model as well as the traffic monitoring algorithms via test-bed experiments under different weather as well as network conditions and show that our proposed hybrid edge-cloud solution outperforms both the cloud-only and edge-only solutions.
\end{abstract}

\begin{IEEEkeywords}
Traffic monitoring, incidents detection, edge-computing, video analytic.
\end{IEEEkeywords}
\vspace{-3mm}
\section{Introduction}\label{sec:Introduction}

Closed-Circuit Television (CCTV) cameras have been widely applied in various surveillance systems. 
For example, New Jersey Department of Trasnportation (NJDOT) installs more than 600 cameras to monitor the traffic in the state. 
During the shift to the smart transportation system, IP-based and conventional cameras are used to not only detect but also mitigate traffic congestion on roads \cite{fan2020buildsensys, fan2020deep}.

In the meantime, establishing a smart transportation system which analyzes video data captured by CCTV cameras becomes challenging. Under the current design, the captured video data need to be sent back to a TMC from cameras. While cameras are located at the edge, the TMC can be viewed as a centralized cloud located somewhere deeper in the network. Therefore, forwarding the video data from cameras to the TMC can raise two major challenges: (1) Processing and analyzing the video data in real-time is nearly impossible. The delay of forwarding the video data on the backhaul network may become serious when the network condition is bad (i.e. limited data rate). (2) Continuous capturing and transferring of video data generates a huge and permanent pressure on the network paths to the TMC.

In order to mitigate the pressure on the backhaul network caused by the video data, one naive strategy is buffering data at the edge with cameras or intermediate network nodes, such as switches and routers. However, CCTV cameras in a transportation system need to capture videos 24/7 without a pause. Therefore, the backhaul network is overwhelmed in the foreseeable future. To tackle the problem of transferring tremendous video data over the backhaul network, edge computing with cloudlet \cite{bonomi2011connected, xiang2020edge} is a more feasible solution.

A cloudlet is considered as a small-scale cloud data center (known as a data center in a box or mobile micro-cloud) which supports resource-intensive applications by providing computing resources at the edge of the network \cite{satyanarayanan2009case}. Therefore, the edge computing concept can reduce the communication bandwidth requirement between the sources of data, i.e., cameras, and the TMC. Through utilizing cloudlets with cameras, edge computing allows us to push the video processing power down to the same level of cameras. Under this new design, each camera in the network plays its own role in processing the video data when it is necessary.

In real world scenarios, the network connection quality between TMC and an individual camera varies from time to time and can significantly affect the video quality \cite{njdot-cameras}. Although cloudlet receives high quality video in real time, their computing resources are often limited. Due to this trade-off in both the TMC and cloudlet, no single solution dominates and it is important to design an adaptive system that switches processing between the edge and the TMC to achieve the trade-off. To this end, we propose a two-tier edge computing model for smart traffic system that  utilizes both the computing capabilities at the edge of the network and the TMC.

This work extends our previous conference publication \cite{kiani2018two}. In the following we highlight the new contributions offered in this work compared to \cite{kiani2018two}.
\begin{itemize}
    
    \item We develop two new algorithms for the vehicle speed detection task deployed on top of our two-tier model.
    \item On both congestion and speed detection, we conduct detailed and extensive experiments that take different weather conditions into consideration. The different weather conditions include rainy, sunny, and snowy conditions. In the conference version, all the experiments were performed under the sunny weather condition.
    \item We introduce new metrics which reveal the performance of Edge and Cloud schemes and the advantage of our Hybrid scheme. Moreover, we also provide statistical analysis which supports our conclusion based on empirical results.
\end{itemize}

The rest of the paper is organized as follows. We review the related works in Section \ref{sec:related}. We highlight the overview of our approach in Section \ref{sec:overview}. We present our design of the two-tier edge computing model in Section \ref{sec:Model}. Section \ref{sec:Algorithm} introduces our traffic detection algorithms in details. Sections \ref{sec:setting} and \ref{sec:results} present our evaluation setting and results. Finally, Section \ref{sec:Conclude} concludes the paper.
\vspace{-3mm}
\section{Related Works}\label{sec:related}

In the past few years, cloud computing has been the major choice for different types of services which require vast amount of data processing since the computing power on the cloud outperforms the capability of the stand-alone device. However, as compared with the fast developing speed of computing power, the network bandwidth is still the bottleneck of the cloud-centralized design. Meanwhile, the video data has become the major part (over 60\%) of the transmitted data in the Internet \cite{Cisco:white-paper}. In order to achieve better service performance,
the paradigm of edge computing has been recently introduced to push the computing resources away from the centralized nodes to the edge of the network. In the past few years, a variety of policies and algorithms have been proposed for the edge network architecture. 
Zhao \textit{et al.}~\cite{zhao2016cluster} proposed a cluster content caching structure for cloud radio access networks (C-RANs) to tackle high power consumption and poor QoS for real-time services caused by significant data exchange in both backhaul and fronthaul links.
A hierarchical Mobile Edge Computing (MEC) model designed based on the principle
of LTE-Advanced backhaul network is introduced in~\cite{kiani2017towards} in which the so-called field, shallow, and deep cloudlets are located in three different tiers of the network. A task scheduling scheme for code partitioning over time and the hierarchical cloudlets is also proposed in~\cite{kiani2017optimal}.

One of the typical use cases of edge computing is video analytic services in smart cities where there is a huge number of video sources with different connection status including surveillance systems, vehicles and mobile devices \cite{kiani2018two}. Thus, utilizing all of these sources together through cloud computing is impossible. To address the aforementioned problems of cloud computing, the edge computing paradigm has been suggested by many researchers as an efficient solution~\cite{shi2016edge,shi2016promise,satyanarayanan2017emergence,de2017resource}. 
For instance, \cite{ali2018edge} proposed an edge computing architecture that supports different phase of deep learning based video analysis from edge to cloud. Moreover, \cite{ledakis2018adaptive} presented its edge computing architecture that integrates the serverless execution model.
In this paper, we focus on the traffic detection problem and propose a two-tier edge computing model that improves the traffic detection accuracy by switching between the video processing at the edge and the cloud based on different network and weather conditions. We utilize and combine the hierarchical edge computing architecture with traffic related video analytic. By considering the advantages and challenges of the video processing at the edge as well as the cloud, our proposed system utilizes the high computing power at the cloud when the network condition between the camera and cloud is good, otherwise, it executes the configuration at the edge that requires less computing power. 
It is worth mentioning that Fog and Edge computing are used interchangeably throughout this work.
\vspace{-3mm}
\section{Approach Overview}\label{sec:overview}

Before diving into the details, we highlight an overview of our approach presented in this work.
\begin{enumerate}
    \item We focus on two common tasks in traffic monitoring system; congestion detection, and speed detection. For each task, we develop two algorithms. One requires less computing power and can analyze over 170 frames per second on a normal PC, so it is suitable to be applied at the edge (cloudlets co-located with cameras) where computing resources are limited. The other one is designed to achieve high accuracy and requires a higher computing power as compared to the first algorithm and is deployed at the TMC (cloud). It is worth mentioning that the design of these algorithms considers the challenges in real-world scenarios. For example, the video quality is affected by many factors, such as the locations that cameras can be deployed, the obstacles in the view (e.g., trees and poles), the shadow from buildings and trees, and the reflection from mirror-like surfaces (e.g., some vehicles), etc.
    
    \item On top of these algorithms, we design a two-tier edge computing model which includes the edge level and the cloud level. This model monitors the network conditions between the edge and the TMC, and switches processing accordingly in an adaptive way. Therefore, it can utilize the high computational power at the cloud when the network condition is good (i.e. unlimited data rate). Otherwise, it relies on the cloudlet at the edge to perform video processing when the network condition is bad.
    
    \item We conduct extensive experiments on different weather conditions to evaluate our proposed two-tier model. Although the cloud-only scheme always uses a highly precise algorithm, its performance is severely affected by the limited network bandwidth that causes jumping frames. By relying on different algorithms under different network conditions, our proposed Hybrid strategy outperforms both the edge-only and cloud-only schemes in terms of detection accuracy in both tasks under all weather conditions. Moreover, we conduct statistical analysis based on experimental results and identify the major factors that affect the performance of the edge and cloud algorithms. The statistical results support our conclusion that the two-tier model achieves higher detection accuracy as compared to the individual Edge and Cloud schemes on different weather conditions.
\end{enumerate}
\vspace{-3mm}

\section{Two-Tier Edge Computing Model}\label{sec:Model}

We consider an edge computing based two-tier network model consisting of the cameras, cloudlets, and the cloud. A cloudlet, e.g., a mini computer, is co-located with each camera at the first tier of the network. All the cameras are assumed to be connected to the cloud located at the TMC using the backhaul network. In the proposed two-tier edge computing based model, we apply two algorithms at the edge and the cloud for each task. In addition, we assume that a video processing decision maker is executed at each cloudlet. This decision maker keeps monitoring the network condition between the camera and the cloud. Based on the strategy detailed below, it decides whether to process video data and detect the traffic at the edge or to forward the captured video to the cloud and rely on the traffic detection results generated there. 

Let's consider a camera of interest and its associated cloudlet. We define $q_{e}$ as the quality of the video at the edge and $q_{c}$ as that at the cloud. As the cloudlet is co-located with the camera, the quality of the video at the edge is assumed to be the same as the recorded video at the camera which represents the highest quality. At the edge, the network condition and its requirement when transferring video data are denoted as $c$ and $c_{t}$, respectively. We define $c,~c_{t} \in [0,1]$ while 0 means disconnected and 1 means connected without bottleneck. Therefore, we can get $q_{c} \equiv \alpha \otimes q_{e} \leq q_{e}$ where $\alpha \equiv \frac{\min(c, c_{t})}{c_{t}}$ is the penalty factor on transferred video. \guanxiong{In other words, the video quality at the cloud is worse than that at the edge unless the network condition is good enough to transmit the video (i.e. $c \geq c_{t}$).} Lastly, we define the detection accuracies of the two algorithms as $A_{c}$ and $A_{e}$ which are monotonically nondecreasing functions. For any given video quality $q$, we have $A_{c}(q) > A_{e}(q)$.

Based on the two-tier edge computing model, we propose the following optimization problem to be solved at the video processing decision maker,
\begin{equation}
\underset{\beta_{e}}{\text{max}} ~ (\beta_{e} A_{e}(q_{e}) + (1-\beta_{e})A_{c}(q_c))\label{equ1}
\end{equation}
where binary variable $\beta_{e}$ is generated by the decision maker and indicates the source of current detection result. As shown in this objective function, we could achieve the trade-off between using the TMC or cloudlets. Since the overall goal is to achieve higher detection accuracy, the solution of this optimization problem takes $\beta_{e}=1$ when the video quality at the TMC is too low to make better detection than cloudlet and vice versa.

In order to solve the optimization problem~(\ref{equ1}), functions $A_{e}(\cdot)$ and $A_{c}(\cdot)$ must be known. However, it is difficult
to choose a universal form for the accuracy functions associated with video qualities. As defined earlier, $q_{c}$ is highly network condition dependent. Therefore, we follow an experimental approach to solve problem~(\ref{equ1}). The idea is to measure the accuracy of $A_{c}(q_{c})$ under different network conditions and accordingly approximate the threshold of network condition $\hat{C}$ under which $A_{c}(q_{c}) < A_{e}(q_{e})$. 
Specifically, we record videos at the server side under different data rate limits which are controlled by an installed software. We empirically compare the video quality and the test accuracy on both congestion and speed detection tasks to determine the threshold. The detailed setting and process are presented in Sec \ref{sec:setting}.
The procedure to be followed at the video processing decision maker is given in Algorithm~\ref{alg:2}. We assume that the network condition is given at the decision maker. That is, the condition of the backhaul network is monitored at the TMC and the decision maker is updated with the current network condition periodically.

\begin{algorithm}
\textbf{INPUT:} The threshold of network condition $\hat{C}$\\
\textbf{OUTPUT:} $\beta_{e}$
\begin{algorithmic}[1]
\WHILE{Decision Maker Works}
    \STATE Measure network condition $c$
    \IF {$c < \hat{C}$}
        \STATE Set $\beta_{e}=1$
    \ELSE
        \STATE Set $\beta_{e}=0$
    \ENDIF
\ENDWHILE
\RETURN $\beta_{e}$
\end{algorithmic}
\caption{Tier Selection for Video Processing}
\label{alg:2}
\end{algorithm}
\vspace{-3mm}

\section{Vehicle Speed Estimation and Traffic Congestion Detection Algorithms}
\label{sec:Algorithm}

To analyze the effect of the edge computing, we use vehicle speed estimation and traffic congestion detection as the evaluation tasks.
Vehicle speed is one of the most widely used metrics in traffic analysis systems.
People can estimate their travel time, and optimize the route based on the speeds of traffic.
Therefore, estimating the speeds of vehicles is a very important task in traffic condition analysis.
Another very common phenomenon in modern cities is the traffic congestion.  
With the increase in the number of vehicles in recent years, urban traffic congestion has also happened more frequently, especially on narrow roads. Traffic congestion detection also becomes a very important task in intellegent transportation systems.
Many video based techniques are proposed to estimate the vehicle speed or to detect the traffic congestion.
In \cite{Luvizon'16}, Luvizon \textit{et al.} presented a vehicle speed measurement method, which analyzes the traffic videos with a spacial resolution of $1920 \times 1080$. 
Famouri \textit{et al.}~\cite{Famouri'18} proposed a vehicle speed estimation method based on motion plane, which works with the videos with a spacial resolution of $960 \times 540$.

\begin{table*}[h]
	\caption{The comparison between some widely used foreground detection methods using the NJDOT traffic surveillance video data set \cite{Hang'18icpr}, \cite{Hang'18mldm}.}
	\centering
	\begin{tabular}{c|c|c|c|c|c}
	\hline \hline
		\textbf{method} & \textbf{GMM} \cite{StaufferGrimson'99} & \textbf{Zivkovic's} \cite{Zivkovic'04}, \cite{Zivkovic'06} & \textbf{Hayman and Eklundh's} \cite{Hayman'03} & \textbf{PAWCS} \cite{St-Charles'16} & \textbf{GFM} \cite{Hang'18icpr}, \cite{Hang'18mldm} \\
		\hline
		Precision & 54\% & 59\% & 83\% & 68\% & 89\% \\
		\hline
		Recall & 86\% & 82\% & 74\% & 83\% & 82\%\\
		\hline
		F-measure & 66\% & 69\% & 78\% & 74\% & \textbf{85\%}\\
		\hline		\hline
		
	\end{tabular}
	
	\label{table:fg}
\vspace{-5mm}
\end{table*}

Most of these methods can achieve a processing speed of 5-15 frames per second or fps on a normal PC. 
However, the cloudlet has a limited computational power on the edge, which makes these methods hard to achieve real time processing.
Additionally, the resolutions they need are too high to be achieved by real world traffic surveillance systems.
Currently, GPU is becoming a common device in normal PCs as deep neural networks are popular. 
Some neural network based methods are proposed for traffic video analysis.
Ke \textit{et al.}~\cite{Ke'18} introduced a traffic congestion detection method using a convolutional neural network(CNN), which can detect the traffic congestion in $1280 \times 720$ videos.
Liu \textit{et al.}~\cite{Liu'20} proposed a detection based speed estimation method that uses neural network models to detect the vehicles and then estimate the speed. 
However, the price for a GPU is still too expensive for real world deployment. 
For some edge devices, such as single-chip microcomputers, the computational power is much less than a desktop PC, these neural network based methods can hardly process the videos in real time. 
Therefore, we develop a statistical based traffic analysis algorithm that can work with our proposed edge computing model without a GPU acceleration.
In comparison, our proposed traffic analysis algorithm can process videos with low resolutions, such as $320 \times 180$, and is able to analyze over $170$ frames per second.

\begin{figure}[tb]
	\begin{center}
		\includegraphics[width=3in]{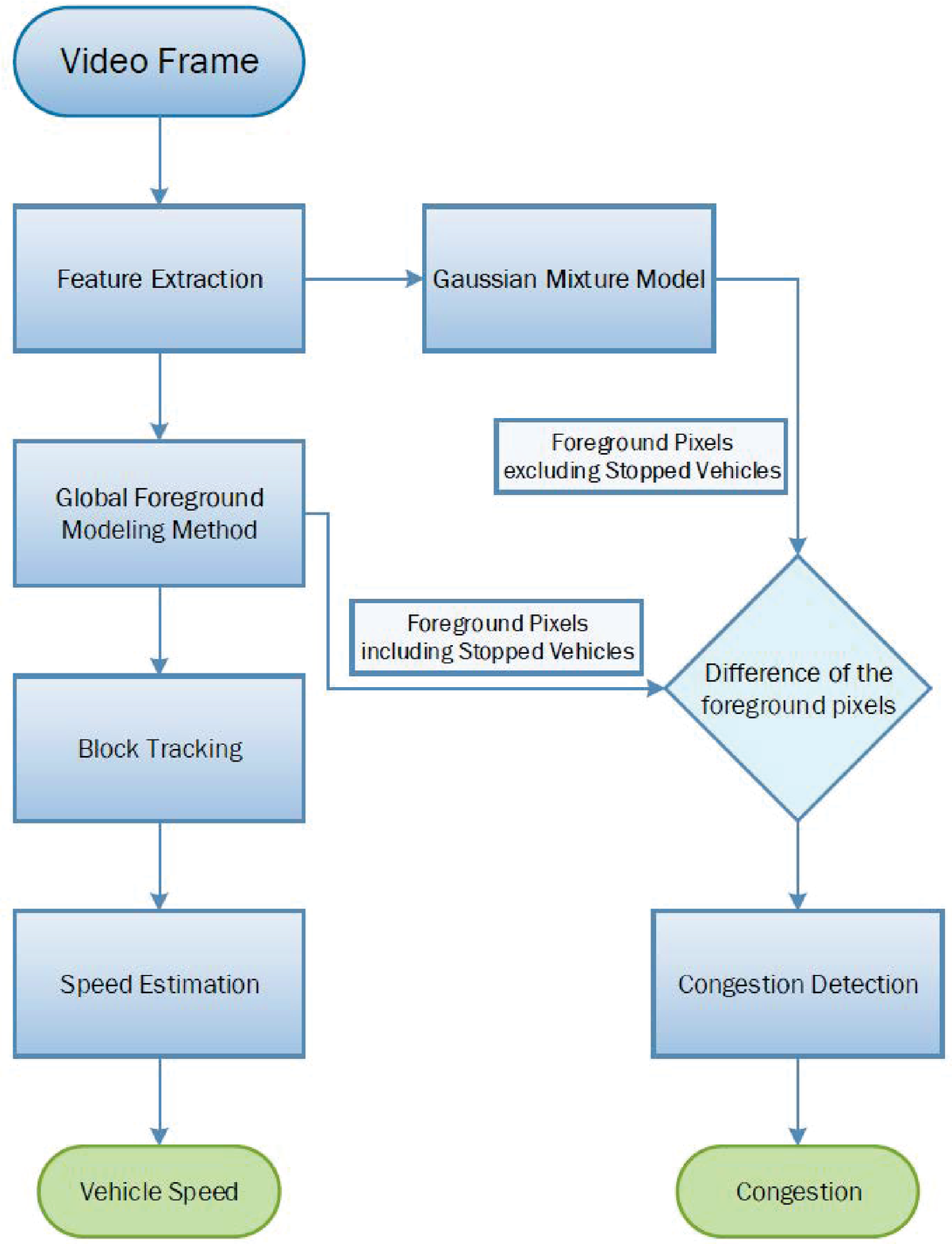}
		\caption{Traffic Analysis Algorithm Work Flow}
		\label{fig:workflow}
	\end{center}
\vspace{-6mm}
\end{figure}

We apply our algorithm with different computational complexities at the edge and the cloud, respectively, by using different resolutions of the input video frames and using feature vectors with different dimensions. The work flow of our traffic analysis algorithm is shown in Fig.~\ref{fig:workflow}.

\subsection{The Foreground Object Detection Method}
\label{fg}

In order to detect the traffic conditions, we first extract the foreground objects from the original frame.
Several methods have been proposed to detect the foreground \cite{Zivkovic'04}, \cite{Hayman'03}, \cite{StaufferGrimson'99}, \cite{St-Charles'16}, \cite{Bouwmans'17}, \cite{Hang'18icpr}, \cite{Hang'18mldm}, among which the Gaussian Mixture Model (GMM) is widely used to detect the foreground due to its efficiency and accuracy. 
One disadvantage of most of the GMM methods; however, is that they cannot detect a foreground object that temporarily stops moving. Actually, many vehicles stop moving during traffic congestion. 
Based on the foreground detection accuracy shown in Table .~\ref{table:fg} and the ability that the Global Foreground Modeling (GFM) method can detect the temporarily stopped vehicles, we therefore apply the GFM method for foreground detection in our edge computing model. 
There are three major steps in the foreground detection method.

First, a background model is built using the Gaussian Mixture Model (GMM) \cite{Zivkovic'04}, \cite{Hayman'03}, \cite{StaufferGrimson'99}.
We estimate the probability density function of the pixel at location ($i,j$) using several Guassian density functions as follows \cite{StaufferGrimson'99}:
\begin{equation}\label{eq:GMM}
p(\mathbf{x}) = \sum_{k=1}^{K}w_kN(\mathbf{M}_k, \Sigma_k)
\end{equation}
where $\mathbf{x} \in \mathbb{R}^d$ indicates the input feature vector in $d$ dimensions, $K$ is the number of Gaussian distributions in the GMM model, $N(\mathbf{M}_k,\Sigma_k)$ is the $k_{th}$ Gaussian distribution with $\mathbf{M}_k$ being the mean vector and $\Sigma_k$ being the covariance matrix, and $w_k$ being the weight associated with $N(\mathbf{M}_k,\Sigma_k)$.
Then, we select the most significant Gaussian density at each location that has the largest weight as the conditional background probability density function $p_{ij}(\mathbf{x}|\omega_b)$ at the pixel located at location ($i,j$).

Second, we build the foreground model using the Global Foreground Modeling (GFM) method \cite{Hang'18icpr}, \cite{Hang'18mldm}.
In the GFM model, we use a series of Gaussian density functions to model all the foreground information; this is different from the background model that builds the model at each location in the frame.
We define $L$ components {$\omega_1$, $\omega_2$, $\cdots$, $\omega_L$} in the GFM model, and each component is associated with a Gaussian density function { $p(\mathbf{x}|\omega_1)$, $p(\mathbf{x}|\omega_2)$, $\cdots$, $p(\mathbf{x}|\omega_L)$}.
$w(\omega_i)$ is the weight for the $i$th component, where $i \in [1,L]$.
Then, for each location, we classify the input feature vector into one Gaussian density function using the Bayes decision rule for minimum error \cite{webb'03}:
\begin{equation} \label{eq:fg}
p(\mathbf{x}|\omega_f)w(\omega_f) = \max_{i=1}^{L} \{p(\mathbf{x}|\omega_i)w(\omega_i)\}
\end{equation}
Thus, we can get the foreground conditional probability density function $p(\mathbf{x}|\omega_f)$ for each location.

Third, we apply the Bayes classifier to classify each pixel into the background class or the foreground class with the conditional probability density functions estimated using the GMM and GFM method.
The prior probability for the background can be estimated using the weight of the first Gaussian distribution in the GMM model, known as $P(\omega_b)=w_1$.
The prior probability for the foreground, $P(\omega_f)$, is estimated as $1-\alpha_1$. 
Location $(i,j)$ will be classified as foreground if the following criterion is satisfied:
\begin{equation}\label{eq:bayes}
p_{ij}(\mathbf{x}|\omega_f)P_{ij}(\omega_f) > p_{ij}(\mathbf{x}|\omega_b)P_{ij}(\omega_b)
\end{equation}

\begin{figure}[tb]
\centering
\begin{minipage}[c]{.5\textwidth}
    \begin{minipage}[c]{.45\textwidth}
    \centering
        \subfloat[Raw Video Frame]{
            \includegraphics[width=\linewidth]{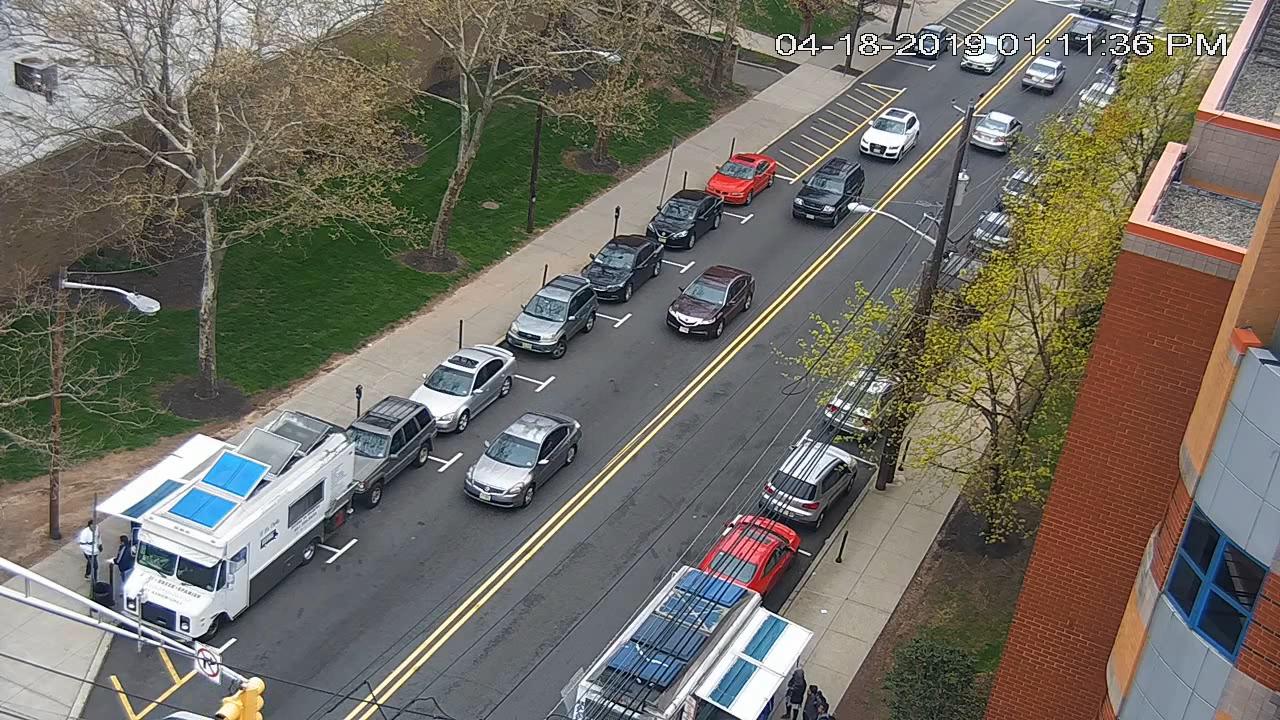}}
    \end{minipage}
    \begin{minipage}[c]{.45\textwidth}
    \centering
        \subfloat[Detected Foreground Mask]{
            \includegraphics[width=\linewidth]{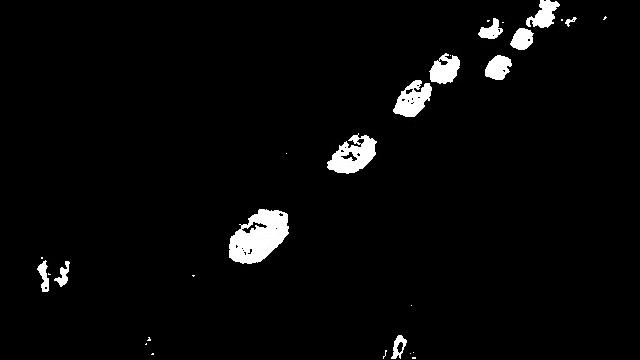}}
    \end{minipage}\vspace{-2mm}
    \caption{Left: A video frame from the traffic surveillance video. Right: The foreground mask detected by the foreground detection method.}
	\label{fig:mask}
\end{minipage}
\vspace{-4mm}
\end{figure}

Therefore, we can get a binary foreground mask for each frame, as shown in Fig.~\ref{fig:mask}.

\hang{
\begin{figure}[!th]
	\begin{center}
		\includegraphics[width=3in]{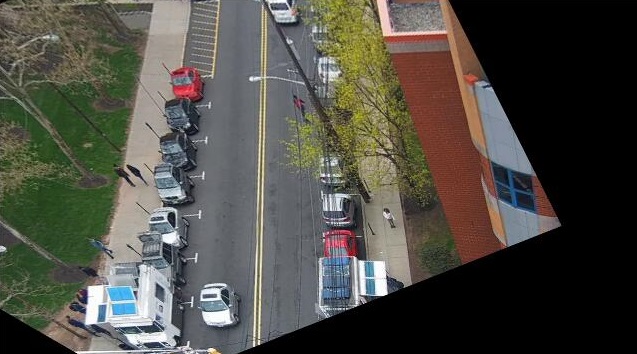}
		\caption{A warped video frame.}
		\label{fig:warped}
	\end{center}
\vspace{-5mm}
\end{figure}
}

\vspace{-7mm}
\subsection{The Vehicle Speed Detection Algorithm}
\label{sec:Speed}

To detect the speed of each vehicle in the traffic, we apply a vehicle speed detection algorithm.
Specifically, we first warp each frame to a top-down view as shown in Fig.~\ref{fig:warped}, in order to eliminate the influence of the viewing-angle.
We then detect each foreground pixel in a warped frame to identify different foreground objects using the foreground object detection method introduced above. 
After that, We link the objects across frames to find the trajectories of the vehicles.
Finally, we use the size of the parking spots as the references to calculate the average speed of the vehicle.
The size of the parking spot is manually measured using a physical scalar.
In the cases that there are no parking spots on the street, we can choose some other factors as the reference object, such as the street marks, etc.
The speed $V$ of each vehicle is estimated as follows:

\begin{equation} \label{eq:speed}
V=c\frac{\sum_{i=1}^{f} \frac{d_{i}}{\Delta t}}{f}
\end{equation}
where $d_{i}$ is the distance that the vehicle has moved between the two adjacent frames, $\Delta t$ is the time difference between two adjacent frames, $f$ is the number of frames the vehicle needs in order to cross two referential lines, and $c$ is a constant used to convert the speed unit to miles per hour or mph. 

The pseudo-code of the vehicle speed detection algorithm is shown in Algorithm.~\ref{alg1}.

\begin{algorithm}
	\caption{Vehicle Speed Detection}
	\label{alg1}
	\textbf{INPUT:} A video \\
	\textbf{OUTPUT:} Speeds of all the vehicles in the video $S$
	
	\begin{algorithmic}[1]
		\FORALL {frames in the video}
			\STATE Value at location $(x,y)$ is $P_{x,y}$
			\FORALL { $P_{x,y} == 1$ }
					\STATE Label the pixel at location $(x,y)$ is $l_{x,y}$
				\IF{no labled pixel adjacent to $(x,y)$}
					\STATE $l_{x,y} = l_{new}$
				\ELSE
					\STATE $l_{x,y} = l_{adjacent}$ 
				\ENDIF
			\ENDFOR
			\STATE The set of labels $L$, the set of trajectories $T$
			\FORALL {$l \in L$}
				\IF{find corresponding trajectory $t \in T$}
					\STATE $l \to t$
				\ELSE
					\STATE $l \to t_{new}$, $t_{new} \to T$
				\ENDIF
			\ENDFOR			
		\ENDFOR
		\FORALL {$t \in T$}
			\STATE Calculate $Speed_{ppf}$ 			
			\STATE $ Speed_{ppf} \to Speed_{mph}, Speed_{mph} \to {S}$
		\ENDFOR	
		\RETURN ${S}$
		
	\end{algorithmic}
\end{algorithm}

\subsection{The Traffic Congestion Detection Algorithm}
\label{sec:Congsetion}

Another traffic condition we used as an evaluation task is traffic congestion. Congestion is a very common incident in transportation systems. We consider that a congestion happens when the vehicles are stopped for a specific amount of time.
This time threshold can be a flexible threshold set up by the user. In this paper, we use 10 seconds as the threshold. 
We manually select the road regions as the region of interest (ROI) in the congestion detection. The congestion happens at the moment that the majority of the road is occupied by stopped vehicles.

\textbf{Definition~\ref{sec:Congsetion}} \textit{Throughout this paper, we consider that a congestion happens when there are more than three vehicles stopped in the ROI and this situation lasts for more than 10 seconds.}

The GFM foreground detection method \cite{Hang'18icpr} \cite{Hang'18mldm} has the capability to correctly detect the temporarily stopped foreground objects, which cannot be achieved by other popular methods \cite{StaufferGrimson'99}, \cite{Hayman'03}, \cite{Zivkovic'04}.
We use this advantage to detect stopped vehicles by comparing the foreground masks obtained by two methods.
Therefore, when the amount of stopped vehicles and their corresponding durations reach the thresholds, we can detect traffic congestion.

The pseudo-code of the congestion detection algorithm is given in Algorithm.~\ref{alg2}.

\begin{algorithm}
	\caption{Traffic Congestion Detection}
	\label{alg2}

		\textbf{INPUT:} A video \\
		\textbf{OUTPUT:} Traffic congestion condition
		
		\begin{algorithmic}[1]
			\STATE Congestion duration $T_c=0$
			\FORALL {frames in the video}
				\STATE Find GFM foreground mask $M_g$ \cite{Hang'18icpr}, \cite{Hang'18mldm}, and Zivkovic foreground mask $M_z$ \cite{Zivkovic'04}, \cite{Zivkovic'06} 
				\STATE Congestion area $A_c=0$
				\FORALL {$(x,y) \in ROI$}
					\STATE Value at location $(x,y)$ in $M_g$ is $G_{x,y}$, in $M_z$ is $Z_{x,y}$
					\IF{$G_{x,y} == 1 \&\& Z_{x,y} == 0 $}
						\STATE $A_c++$
					\ENDIF
				\ENDFOR
				\STATE Threshold for congestion area $\tau_a$
				\IF {$A_c >\tau_a$}
					\STATE $T_c++$
				\ELSE
					\STATE $T_c--$
				\ENDIF
				\STATE Threshold for congestion time $\tau_t$
				\IF {$T_c >\tau_t$}
					\RETURN true
				\ELSE
					\RETURN false
				\ENDIF
			\ENDFOR

	\end{algorithmic}
\end{algorithm}

\vspace{-6mm}
\subsection{Different Configurations for Edge Computing and Cloud Computing}
\label{config}

In the two-tier edge computing model, we apply our vehicle speed detection and traffic congestion detection algorithm using two different configurations at the edge and in the cloud, which require different computation power and can achieve different accuracy levels.

\textit{\textbf{Configuration-1}}. This configuration is used in the cloud, which has sufficient computational power. 
We use the 3 dimensional RGB color values as the input feature vector for the foreground detection method and scale each frame to $640 \times 360$. 
Additionally, we apply the morphological processes to enhance the foreground detection results.
The processing speed of our algorithm can be up to 22 fps on the mini PC used in the experiment, which is faster than the 15 fps of the original video. 

\textit{\textbf{Configuration-2}}. This configuration is used at the edge, which has limited computational power. 
We use the intensity value as the input feature for our algorithm and resize each frame to the spatial resolution of $320 \times 180$. 
Under this configuration, the running speed is increased to 173 fps on the same mini PC.

The difference between configuration-1 and configuration-2 is mainly reflected in three aspects.
First, the input feature vectors are different. 
As the three dimensional RGB color values can provide more information than the intensity value of a pixel, the discriminating power of the feature vectors in configuration-1 is stronger than that in configuration-2. 
As a result, the foreground mask detected using configuration-1 is more reliable than the foreground mask detected using configuration-2.
However, higher dimensional feature vectors require more computation when generating the foreground mask, which causes configuration-1 to be slower in speed.
Second, configuration-1 involves the morphological processes.
The morphological processes can help reduce noises and fill holes, but will slow down the processing speed. 
Third, the spatial resolution used in configuration-1 is twice that of configuration-2 in both width and height.
For configuration-2, the real world area represented by one pixel is four times that of the high resolution frames, and so the effect of the noise in the foreground masks is much larger than that of configuration-1.
On the other hand, the number of pixels that needs to be processed in configuration-2 is only one quarter that for configuration-1, which will make it run four times faster in the foreground mask extraction step.
These three major differences in the configuration result in configuration-1 achieving higher accuracy than configuration-2, but at a relatively slower speed (still in real time).

\vspace{-3mm}
\subsection{The Hybrid Computing Strategy}
\label{hybrid}
As we described in Sec.~\ref{config}, the configurations of edge computing and cloud computing have their own advantages as the edge is faster and the cloud is more accurate.
To combine the advantages of these two methods, we introduce a hybrid computing strategy.
We set up a threshold to determine the status of the internet condition (e.g, 300KB/s). This threshold is determined based on experimental evaluations. If the connection speed is over the threshold, we recognize the network condition to be good; otherwise, we consider it to be bad.
When the network condition is good, we can transfer the video data to our cloud computing center, which has enough computation power and process the video using the cloud computing configuration.
When the network condition is bad, we do not transfer the data, but process the video locally using the edge computing configuration.
Thus, we can make sure that at any time the computation power is enough to perform fast processing and the video analysis result is as accurate as possible.

\vspace{-3mm}

\section{Evaluation Setting}\label{sec:setting}

In this section, we present the evaluation setting used in this work. First, we provide the test-bed and the hyper-parameter settings. Then, we detail our experiments and the corresponding analysis.

\begin{figure}[tb]
\centering
\begin{minipage}[c]{.45\textwidth}
\centering
    \includegraphics[width=\linewidth]{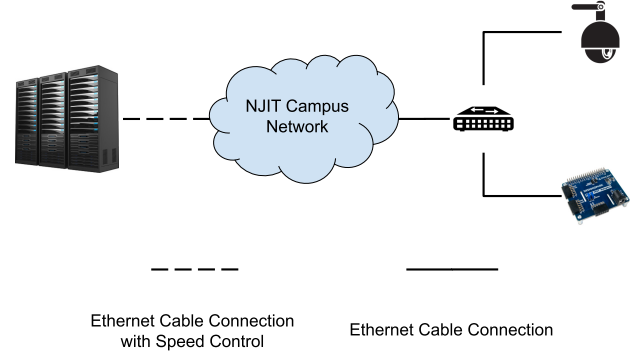}
    \vspace{-5mm}
    \caption{Test-bed Connection}
    \label{fig:connection}
\end{minipage}
\vspace{-4mm}
\end{figure}

\vspace{-3mm}
\subsection{Test-bed and Hyper-parameter}

In order to evaluate the performance, we build a small test-bed of our proposed two-tier edge computing model. The test-bed contains an IP-based camera (Speco o2p12x), a cloudlet mini PC (Intel NUC 7i7bhn) and a cloud server (four Dell poweredge servers). The cloudlet mini PC is equipped with a 2-core 3.5GHz i7-7567U processor, 16G DDR4 RAM and 1TB HDD. On the cloud side, we have two servers with 10-core Intel Xeon E5-2630 v4 2.2GHz processor and the other two are equipped with 8-core Intel Xeon E5-2620 v4 2.1GHz processor. This test-bed is deployed in the New Jersey Institute of Technology (NJIT) campus and the detailed connection is visualized in Fig. \ref{fig:connection}. Since the backhaul network of this test-bed is the NJIT campus network, the network condition between the camera and cloud servers is always good. In order to emulate the real world environment, we utilize the NetLimiter 4 with a predefined script to automatically change the connection speed.

\begin{figure}[tb]
\centering
\begin{minipage}[c]{.4\textwidth}
\centering
    \includegraphics[width=\linewidth]{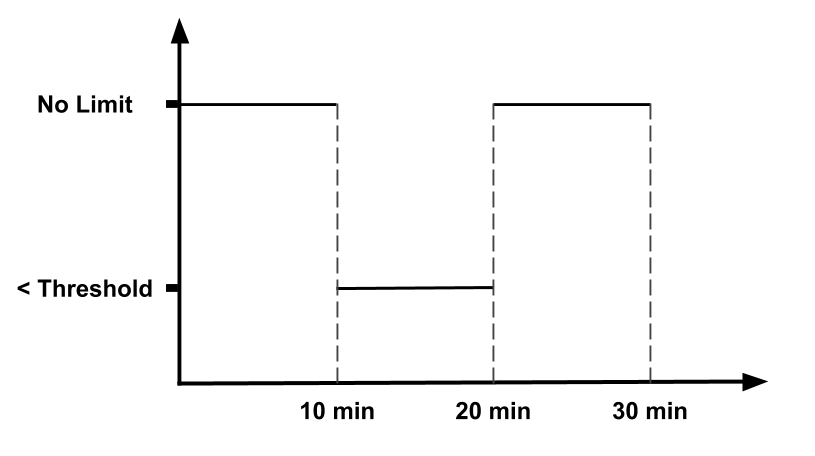}
\end{minipage}
\vspace{-3mm}
\caption{The process of limiting the data rate over time}\label{fig:speed}
\vspace{-3mm}
\end{figure}

\vspace{-3mm}
\subsection{Experiment and Analysis}

Because there is no available public data set, we build our own video data set using an IP-based camera. The IP-based camera in our test-bed is located on the rooftop of the Laurel Hall in the NJIT campus. This camera captures the traffic on the Warren Streen in Newark (New Jersey). To evaluate our proposed edge computing model, the video recorded by the camera is sent to both the local cloudlet and the remote server. As a result, the received video by the server is affected by packet drops due to the limited network bandwidth. 
In the experiments, we follow a strict rule to record videos. Our recording activity is conducted in the normal business days within the time period from 1pm to 4pm. Therefore, all of our recorded videos reflect the similar business day traffic during off-peak hours. 
We record one 30-minute video for each weather condition (e.g., sunny, rainy, or snowy) to make a fair comparison. The different weather conditions are considered to have different lighting conditions and the bad weather conditions may involve noises in the video frames. 
In each experiment, the network connection between the camera and the server is controlled by a pre-defined script.
As shown in Fig. \ref{fig:speed}, the 30-minute record has a middle 10-minute with data rate limit. The network condition is good when there is no data rate limit and is bad when data rate limit is applied.

To capture the performance of our proposed two-tier model in real-world scenarios, we first evaluate the effect of network conditions on the video quality at the server side. In the experiments, we choose to record video at the server side with different data rate limits which include no-limit, 500KB/s, 400KB/s and 300KB/s. From our repeated experiments, we evaluate the quality of recorded videos and compare the test accuracies of congestion and speed detection when they are fed into algorithm $A_{c}$ in Section \ref{sec:Model}. As a validation process, we also compare the selected threshold with the data rate that traffic monitoring system deployed in the real-world scenario can obtain.

Once the threshold on data rate is identified from previous experiments, we start the evaluation of our two-tier edge computing model. In this experiment, our two-tier model (dubbed \textbf{Hybrid}) is compared with edge-only (dubbed \textbf{Edge}) as well as cloud-only (dubbed \textbf{Cloud}) in both congestion and speed detection tasks. The evaluation results of the Cloud scheme setting on good (unlimited data rate) and bad (limited data rate) network conditions are represented as \textbf{Cloud+} and \textbf{Cloud-}, respectively. For Cloud+, the data rate on network connection between the server and the camera is unlimited. For Cloud-, the data rate is limited to the pre-defined threshold, which we identify as 300KB/s. In addition to the switching between good and bad network conditions, we also repeat the recording activity in three different weather conditions (sunny, rainy and snowy) to show their effect on test accuracies of all three settings.

\Lee{To ensure the credibility and objectivity of performance evaluation, we collected ground truth data by manually retrieving the speed of individual vehicles from the video footage used in the experiments. It is worth noting that the data collection was repeated by the same data collector at least three times to minimize potential human measurement errors. The average value of individual vehicle's speed samples was finally used as the ground truth for the evaluation.}
Let $\tilde{x}_{i} (i \in \left\{1, 2, 3, \dots, n\right\})$ be the ground truth at the discretized time stamp $i$ and $x_{i}$ be its corresponding detection result. We define several metrics that are used in evaluation. Firstly, we utilize the detection error rate as the metric for both congestion and speed detection and denote it as $\epsilon_{C}$ and $\epsilon_{S}$ for congestion and speed detection, respectively. Moreover, we also provide the root mean square (RMS) error, $\epsilon_{RMS}$, calculated from speed detection results. Through its comparison with the ground truth of the average speed, we can quantitatively measure the performance of speed detection. The mathematical definitions of these metrics are presented as follows. 
\begin{align}
    & \epsilon_{C} = \frac{\sum_{i=0}^{n} I(\tilde{x}_{i} - x_{i})}{\underset{i}{\Sigma} I(\tilde{x}_{i})}
    & \epsilon_{S} = \frac{1}{n} \sum_{i=0}^{n} \frac{|\tilde{x}_{i} - x_{i}|}{\tilde{x}_{i}} \\
    & \epsilon_{RMS} = \sqrt{\frac{1}{n} \sum_{i=0}^{n} (\tilde{x}_{i} - x_{i})^{2}}
\end{align}
Beyond the presentation of our experimental results, we also conduct statistical analysis on top of the results to reveal major factors that influence the accuracies of congestion and speed detection under different weather conditions.

\begin{table}[tb]
    \centering
    \caption{Summary of abbreviations}\label{table:abbreviations}
    \vspace{-3mm}
    \begin{tabular}{ c | p{6cm} }
     \hline \hline
     \textbf{Abbreviations} & \textbf{Description} \\
     \hline
     Hybrid
     & The proposed scheme that can switch between the edge and the cloud.\\
     \hline
     Edge
     & The scheme that always relies the cloudlets on edge.\\
     \hline
     Cloud
     & The scheme that always relies the server on cloud.\\
     \hline
     Cloud+
     & The cloud-only scheme under the good network condition (i.e. unlimited data rate).\\
     \hline
     Cloud-
     & The cloud-only scheme under the bad network condition (i.e. limited data rate).\\
     \hline
     $\epsilon_{C}$
     & The detection error rate in congestion detection. A lower error corresponding to a better congestion detection performance.\\
     \hline
     $\epsilon_{S}$
     & The detection error rate in speed detection. A lower error corresponding to a better speed detection performance.\\
     \hline
     $\epsilon_{RMS}$
     & The root mean square error that quantitatively measures the performance of speed detection. A lower error corresponding to a better speed detection performance.\\
     \hline
    \end{tabular}
\end{table}

To help the reader understand our experiments, we summarize the abbreviations in Table \ref{table:abbreviations} with intuitive description.
\vspace{-3mm}
\begin{figure*}[tb]
\begin{minipage}[c]{.48\textwidth}
    \begin{minipage}[c]{.48\textwidth}
    \centering
        \subfloat[Congestion Error Rate]{
            \includegraphics[width=\linewidth]{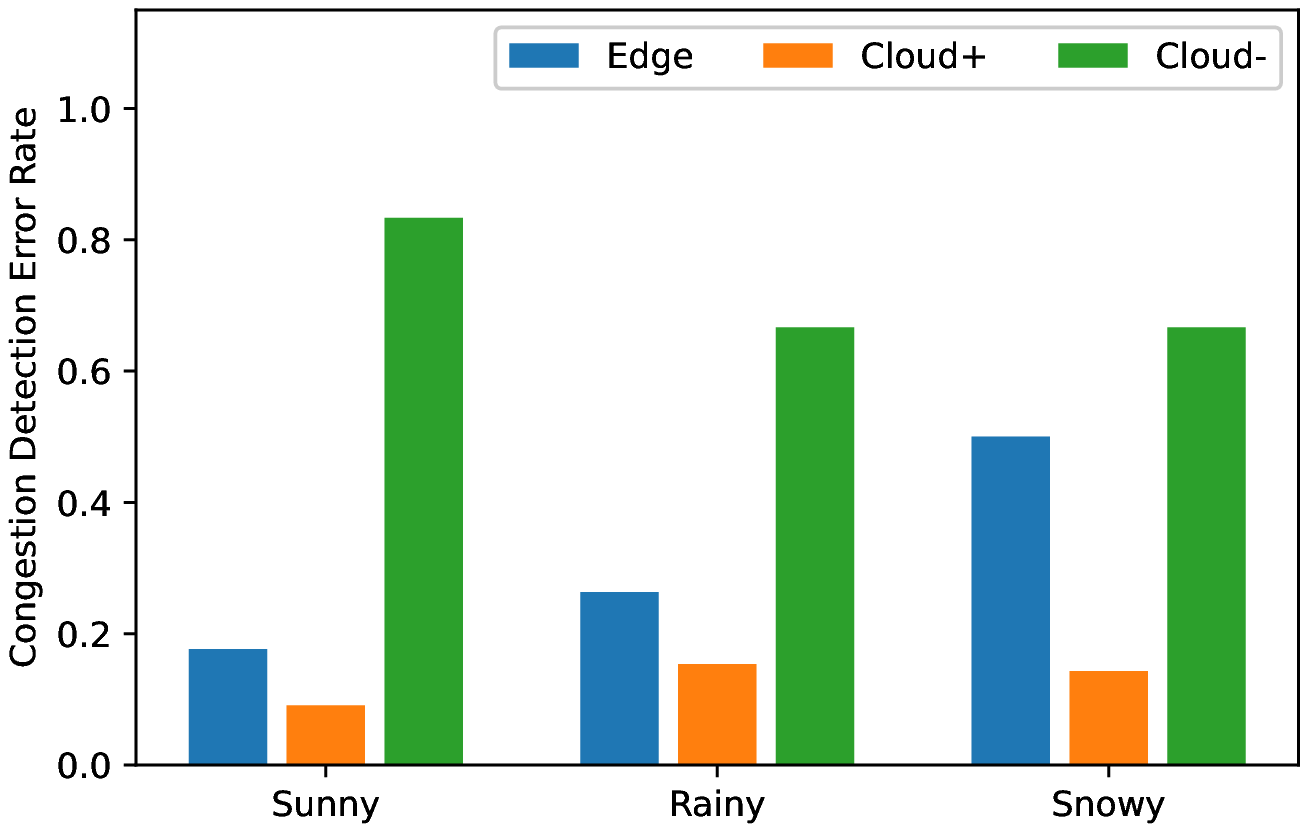}}
    \end{minipage}
    \begin{minipage}[c]{.48\textwidth}
    \centering
    \subfloat[Speed Detection Error Rate]{
        \includegraphics[width=\linewidth]{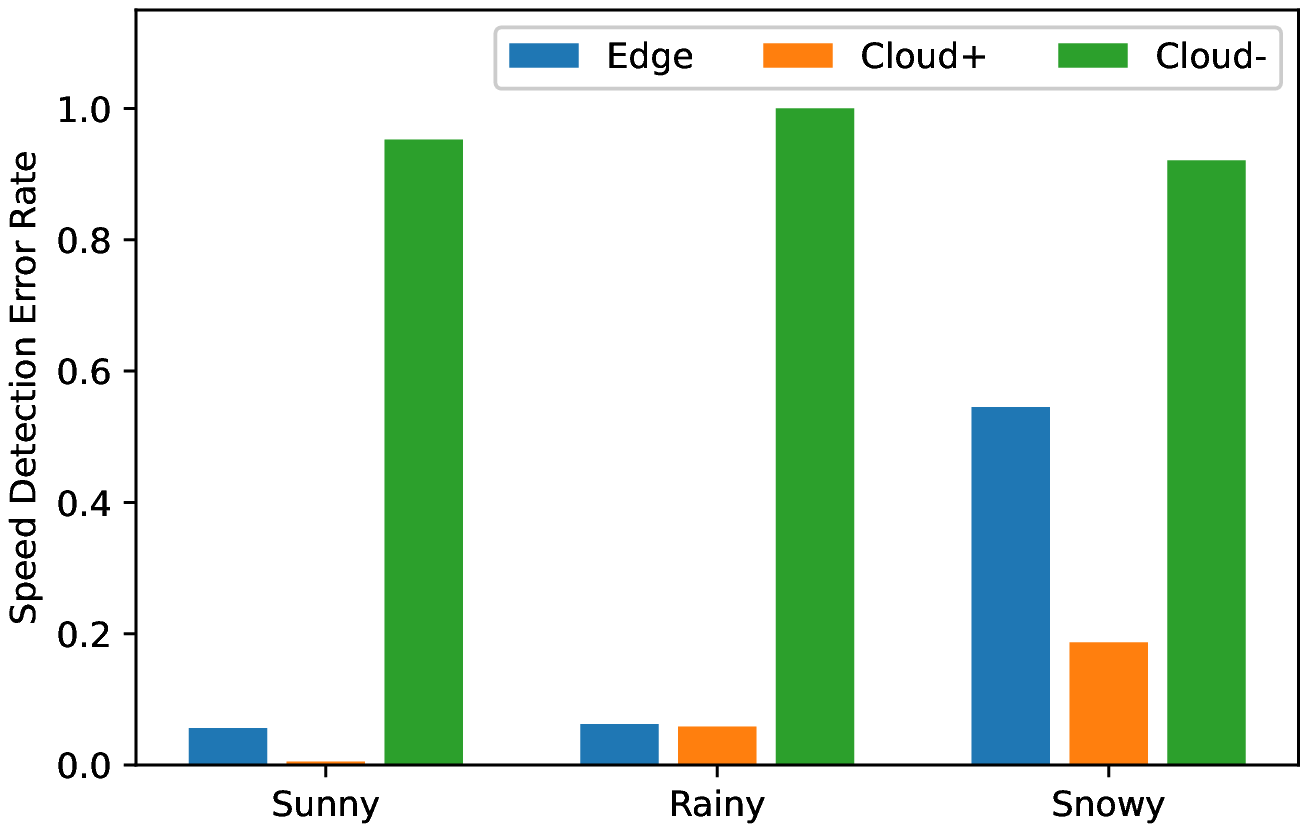}}
    \end{minipage}\vspace{-2mm}
    \captionsetup{width=.8\textwidth}
    \caption{Congestion (Left) and Speed Detection (Right) Error Rate of Edge and Cloud Schemes}
    \label{fig:compare1}
\end{minipage}
\begin{minipage}[c]{.48\textwidth}
    \begin{minipage}[c]{.48\textwidth}
    \centering
    \subfloat[Congestion Error Rate]{
        \includegraphics[width=\linewidth]{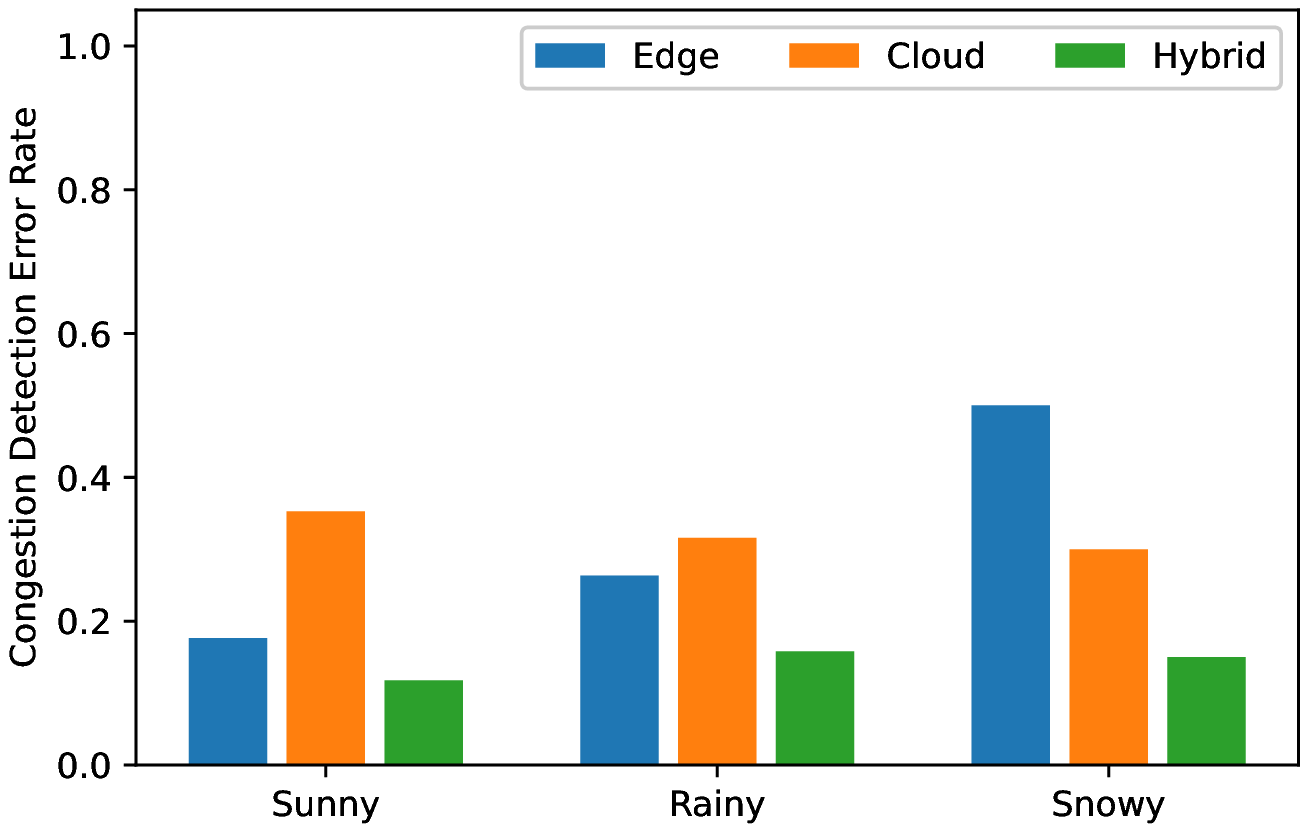}}
    \end{minipage}
    \begin{minipage}[c]{.48\textwidth}
    \centering
    \subfloat[Speed Detection Error Rate]{
        \includegraphics[width=\linewidth]{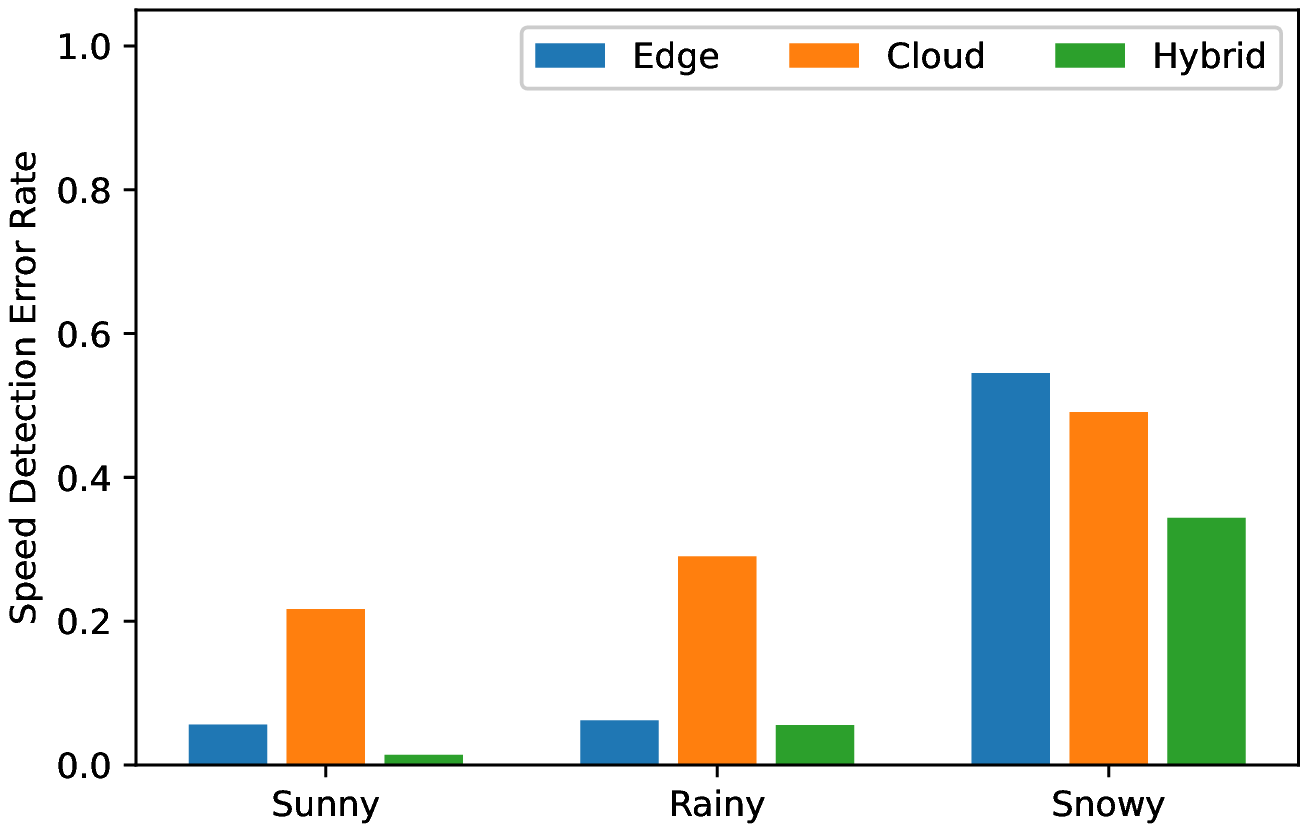}}
    \end{minipage}\vspace{-2mm}
    \captionsetup{width=.9\textwidth}
    \caption{Congestion (Left) and Speed Detection (Right) Error Rate of Edge, Cloud and Hybrid Schemes}
	\label{fig:compare2}
\end{minipage}
\vspace{-5mm}
\end{figure*}

\section{Evaluation Result}\label{sec:results}
\vspace{-2mm}
\subsection{Cloud Scheme under Different Data Rate Limits}
\vspace{-2mm}

The first phase of our evaluation aims at comparing the performance ($\epsilon_{C}$, $\epsilon_{S}$ and $\epsilon_{RMS}$) of the Cloud scheme under different data rate limits. The experiment contains four different data rate limits which include no-limit, 500KB/s, 400KB/s and 300KB/s. As a reference, we also add results of Edge scheme on videos recorded by the cloudlet. We conduct the experiment on the normal weather to eliminate other effects. The experiment results are summarized in Table \ref{table:diff-rate}.

The results clearly show the following: (1) The 500KB/s limit only slightly degenerates the performance of the Cloud scheme and all error metrics are lower than the Edge scheme. (2) The 400KB/s limit causes the Cloud scheme to perform worse than the Edge scheme on speed detection. (3) When we increase the limit to 300KB/s, the Cloud scheme performs worse than the Edge scheme on both congestion and speed detection tasks. These results show that both congestion and speed detection tasks are affected by the connection speed limit and speed detection task is more sensitive. The reason is that the videos incur increasingly more serious jumping frames. These jumping frames significantly corrupt the quality and continuity of videos, and speed detection has higher requirement on continuity than congestion detection.

Based on the results in this evaluation phase, we decide to utilize 300KB/s as the threshold on data rate. It means that the data rate between the server and the camera is less than or equal to 300KB/s under Cloud-. For the Hybrid scheme, when the data rate is more than this value, we perform processing on the cloud, otherwise we utilize the edge. We assume the TMC is equipped with the fastest IEEE Ethernet solution on the market, IEEE P802.3bs, which supports up to 400Gbit/s data rate \cite{eiselt2016first}. Moreover, we also assume that the entire data rate could be shared between the TMC server and different cameras at the edge without any overhead. 
Given that the total bandwidth of the TMC is 400Gbit/s and the good network condition (Cloud+) requires the data rate to be higher than 300KB/s, the TMC can only support no more than 166,666 cameras to operate simultaneously under Cloud+.
As compared with the surveillance network in China (170 million cameras), less than 0.2 million cameras are not even enough to cover the state. Therefore, building the national or state-scaled surveillance network must face and solve the problem of bad network condition between the TMC and some edge cameras.

\begin{table}[tb]
    \caption{Performance of different schemes}
    \vspace{-3mm}
    \label{table:diff-rate}
    \begin{center}
    \begin{tabular}{ c | c  c  c  c  c }
    \hline \hline
    & No-limit & 500KB/s & 400KB/s & 300KB/s & Edge \\
    \hline
    $\epsilon_{C}$ & 9.09\% & 13.33\% & 13.33\% & 83.33\% & 17.65\% \\
    $\epsilon_{S}$ & 0.53\% & 4.00\% & 63.16\% & 95.24\% & 5.60\% \\
    $\epsilon_{RMS}$ & 0.3259 & 3.1676 & 14.5611 & 19.5267 & 1.1145 \\
    \hline
    Avg Speed & 17.0514 & 17.5033 & 17.6075 & 19.5656 & 17.6131 \\
    \hline \hline
    \end{tabular}
    \end{center}
\vspace{-3mm}
\end{table}

\begin{table}[tb]
\begin{center}
\caption{Comparison of Hybrid, Edge and Cloud Schemes}\label{tab:error-rate}
\begin{tabular}{ c  c | c  c  c }
    \hline \hline
    &  & Sunny & Rainy & Snowy \\
    \hline
    \multirow{5}{*}{\begin{sideways}$\epsilon_{C}$\end{sideways}} & Edge & 17.65\% & 26.32\% & 50.00\% \\
    & Cloud & 35.29\% & 31.58\% & 30.00\% \\
    & Cloud+ & 9.09\% & 15.38\% & 14.29\% \\
    & Cloud- & 83.33\% & 66.67\% & 66.67\% \\
    & Hybrid & 11.76\% & 15.79\% & 15.00\% \\
    \hline
    \multirow{10}{*}{\begin{sideways}$\epsilon_{S}$\end{sideways}} 
    & \multirow{2}{*}{Edge} & 5.60\% & 6.21\% & 54.50\% \\
    & & $\pm$ 0.62\% & $\pm$ 1.53\% & $\pm$ 16.02\% \\
    & \multirow{2}{*}{Cloud} & 21.69\% & 29.03\% & 49.07\% \\
    & & $\pm$ 8.28\% & $\pm$ 10.03\% & $\pm$ 17.31\% \\
    & \multirow{2}{*}{Cloud+} & 0.53\% & 5.85\% & 18.69\% \\
    & & $\pm$ 0.36\% & $\pm$ 1.79\% & $\pm$ 15.57\% \\
    & \multirow{2}{*}{Cloud-} & 95.24\% & 100.00\% & 92.10\% \\
    & & $\pm$ 9.33\% & $\pm$ 0.00\% & $\pm$ 16.49\% \\
    & \multirow{2}{*}{Hybrid} & 1.41\% & 5.56\% & 34.37\% \\
    & & $\pm$ 0.52\% & $\pm$ 1.41\% & $\pm$ 14.96\% \\
    \hline
    \multirow{5}{*}{\begin{sideways}$\epsilon_{RMS}$\end{sideways}} & Edge & 1.1145 & 1.4644 & 2.5474 \\
    & Cloud & 9.2339 & 9.7376 & 2.1638 \\
    & Cloud+ & 0.3259 & 1.3871 & 1.4486 \\
    & Cloud- & 19.5267 & 19.4761 & 2.8882 \\
    & Hybrid & 0.5467 & 1.3471 & 1.6950 \\
    \multicolumn{2}{c |}{Avg Speed} & 17.6131 & 18.0937 & 3.3756 \\
    \hline \hline
\end{tabular}
\end{center}
\vspace{-3mm}
\end{table}

\begin{figure*}[tb]
\centering
\begin{minipage}[c]{.32\textwidth}
    \centering
    \subfloat[$\epsilon_{C}$ in Sunny Day]{
        \includegraphics[width=\linewidth]{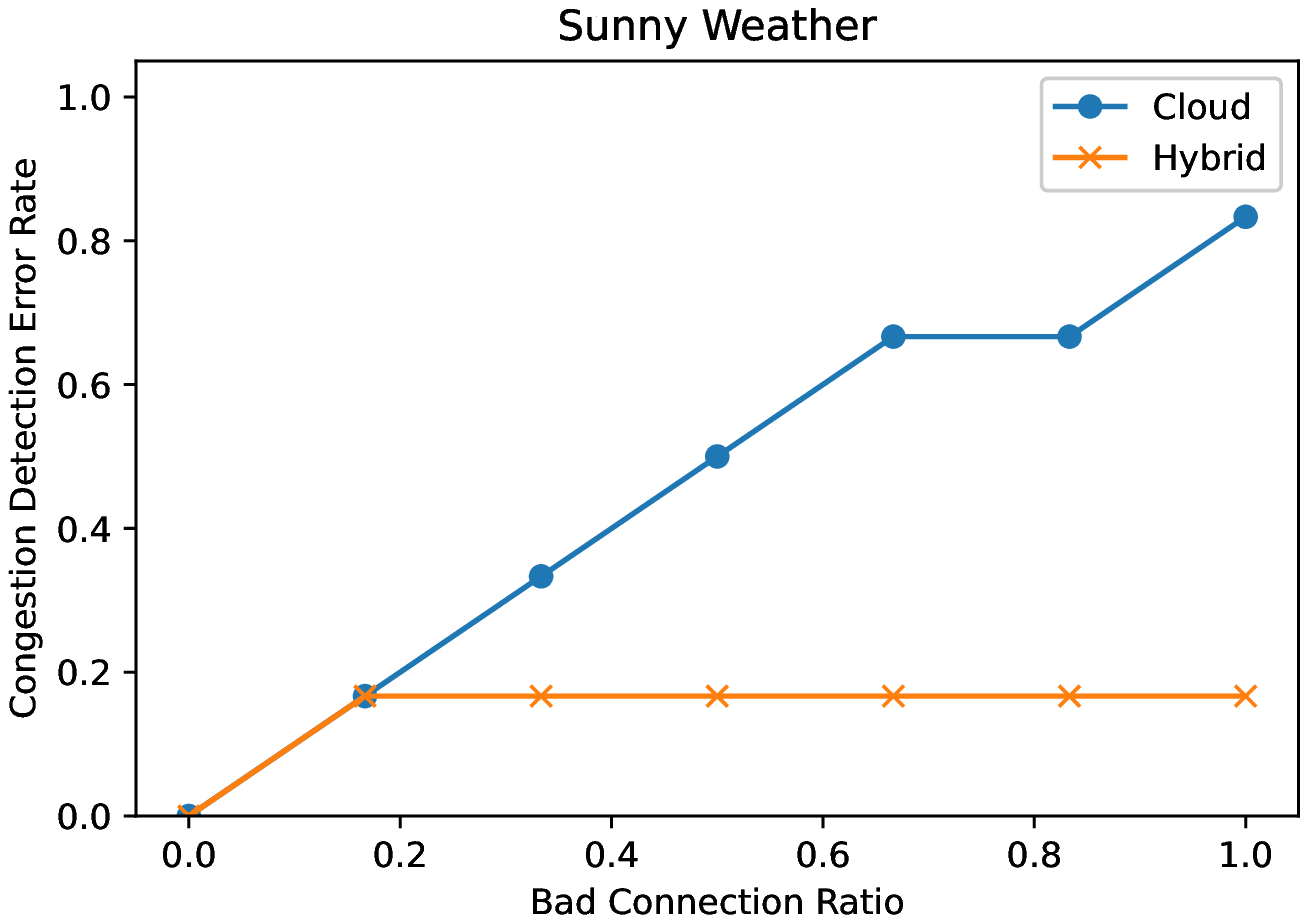}}
\end{minipage}
\begin{minipage}[c]{.32\textwidth}
    \centering
    \subfloat[$\epsilon_{C}$ in Rainy Day]{
        \includegraphics[width=\linewidth]{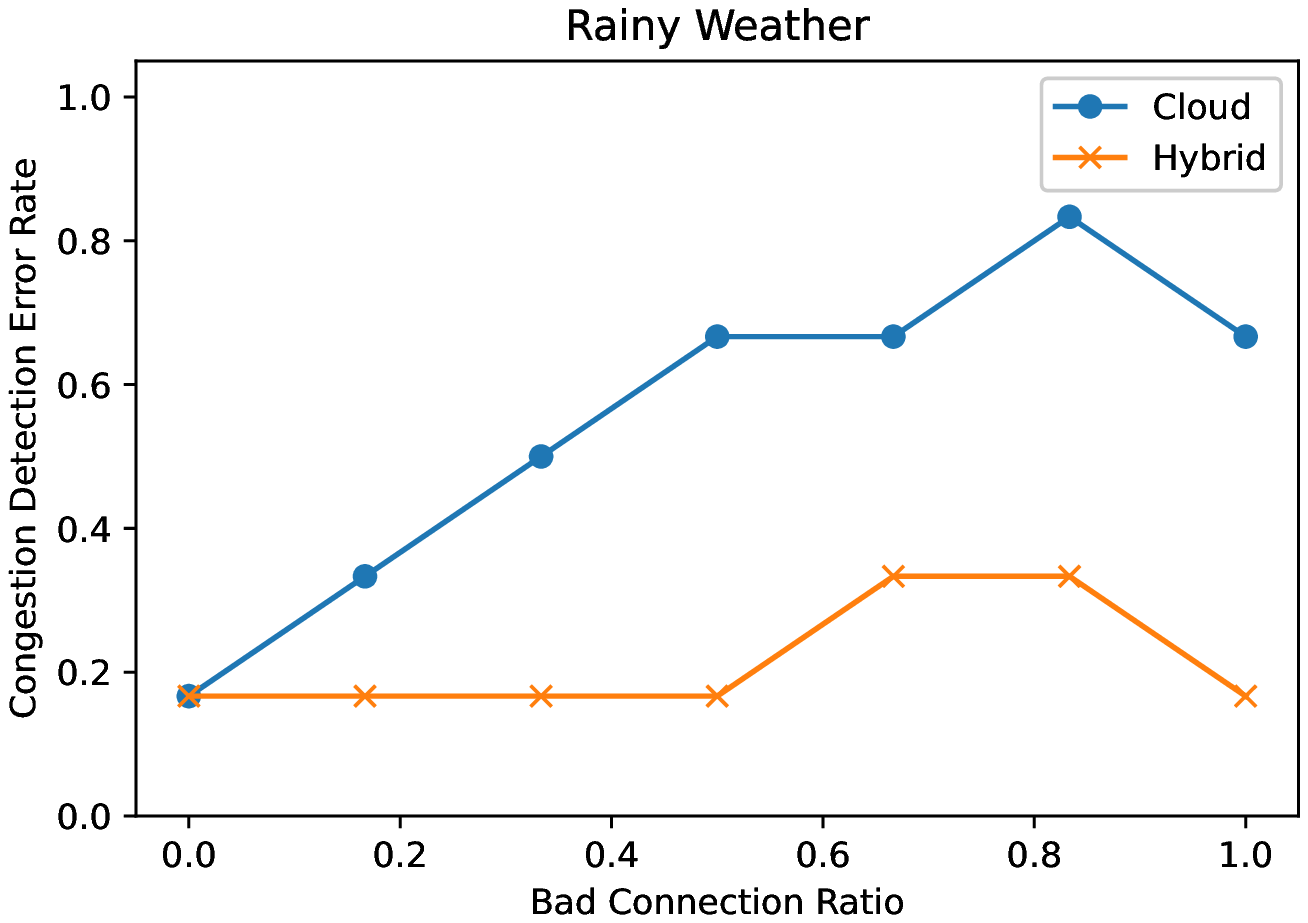}}
\end{minipage}
\begin{minipage}[c]{.32\textwidth}
    \centering
    \subfloat[$\epsilon_{C}$ in Snowy Day]{
        \includegraphics[width=\linewidth]{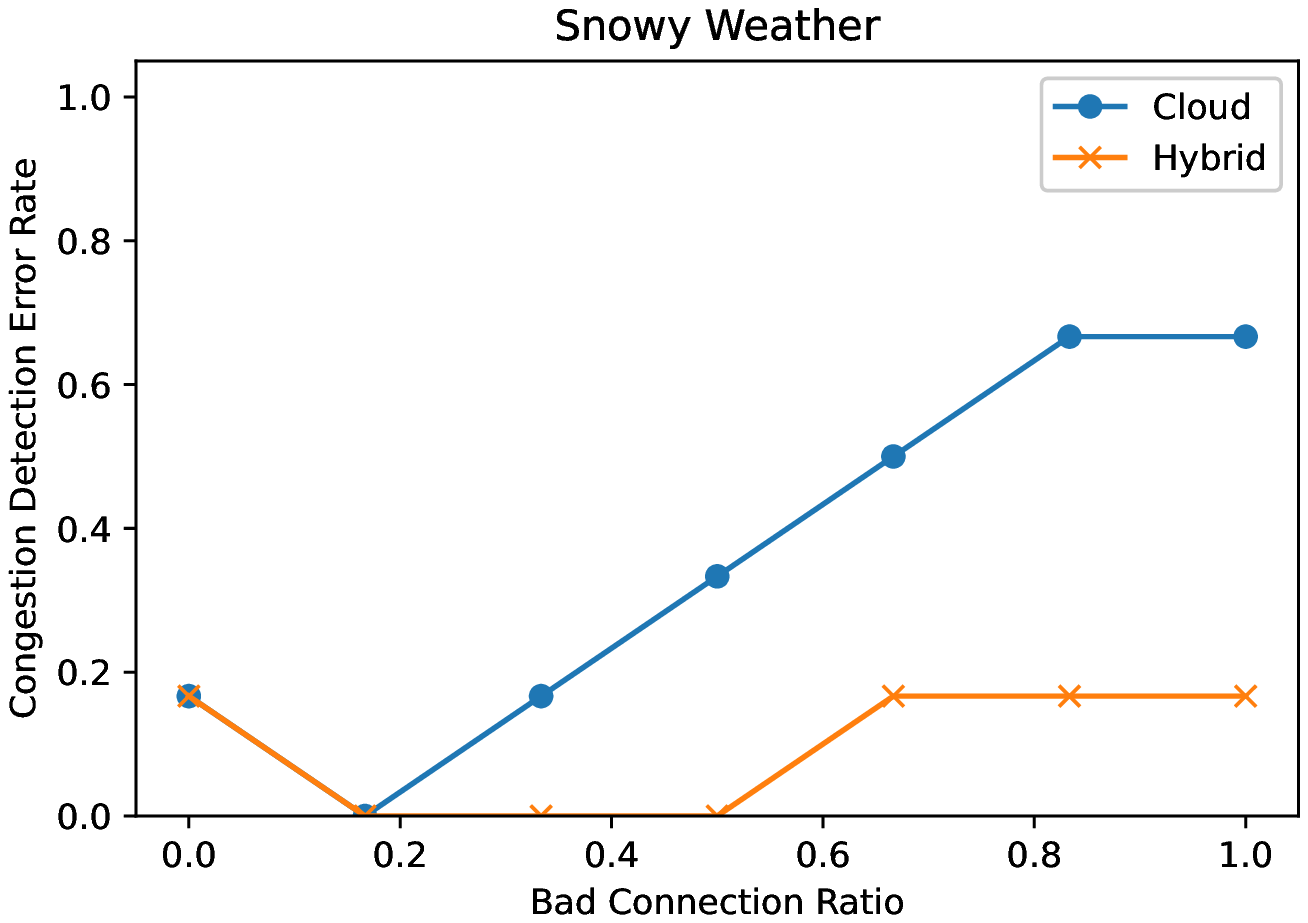}}
\end{minipage}
\vspace{-2mm}
\caption{$\epsilon_{C}$ under Different percentage of time for bad network condition}
\vspace{-5mm}
\label{fig:congestion-slide}
\end{figure*}
\begin{figure*}[tb]
\centering
\begin{minipage}[c]{.32\textwidth}
    \centering
    \subfloat[$\epsilon_{S}$ in Sunny Day]{
        \includegraphics[width=\linewidth]{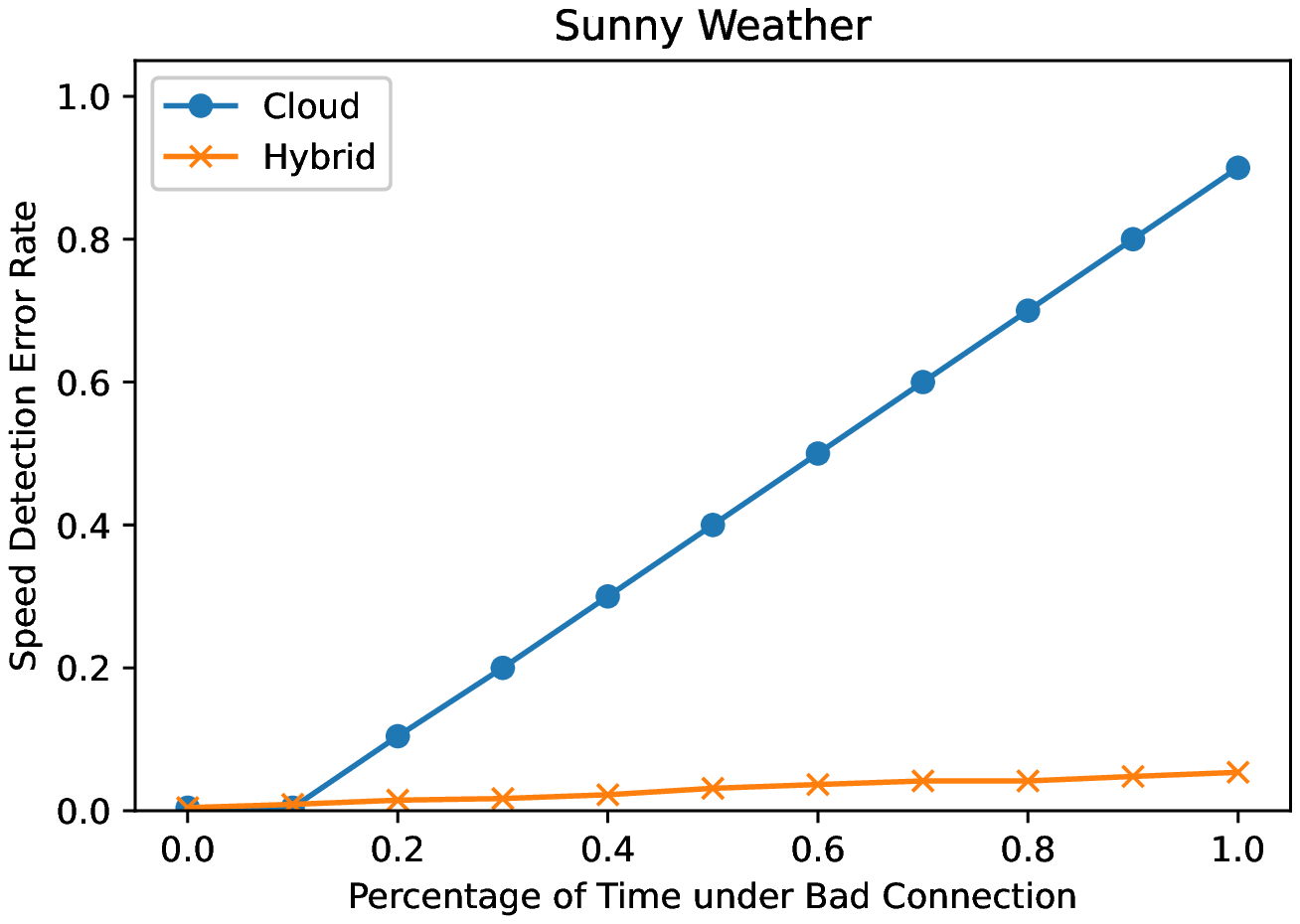}}
\end{minipage}
\begin{minipage}[c]{.32\textwidth}
    \centering
    \subfloat[$\epsilon_{S}$ in Rainy Day]{
        \includegraphics[width=\linewidth]{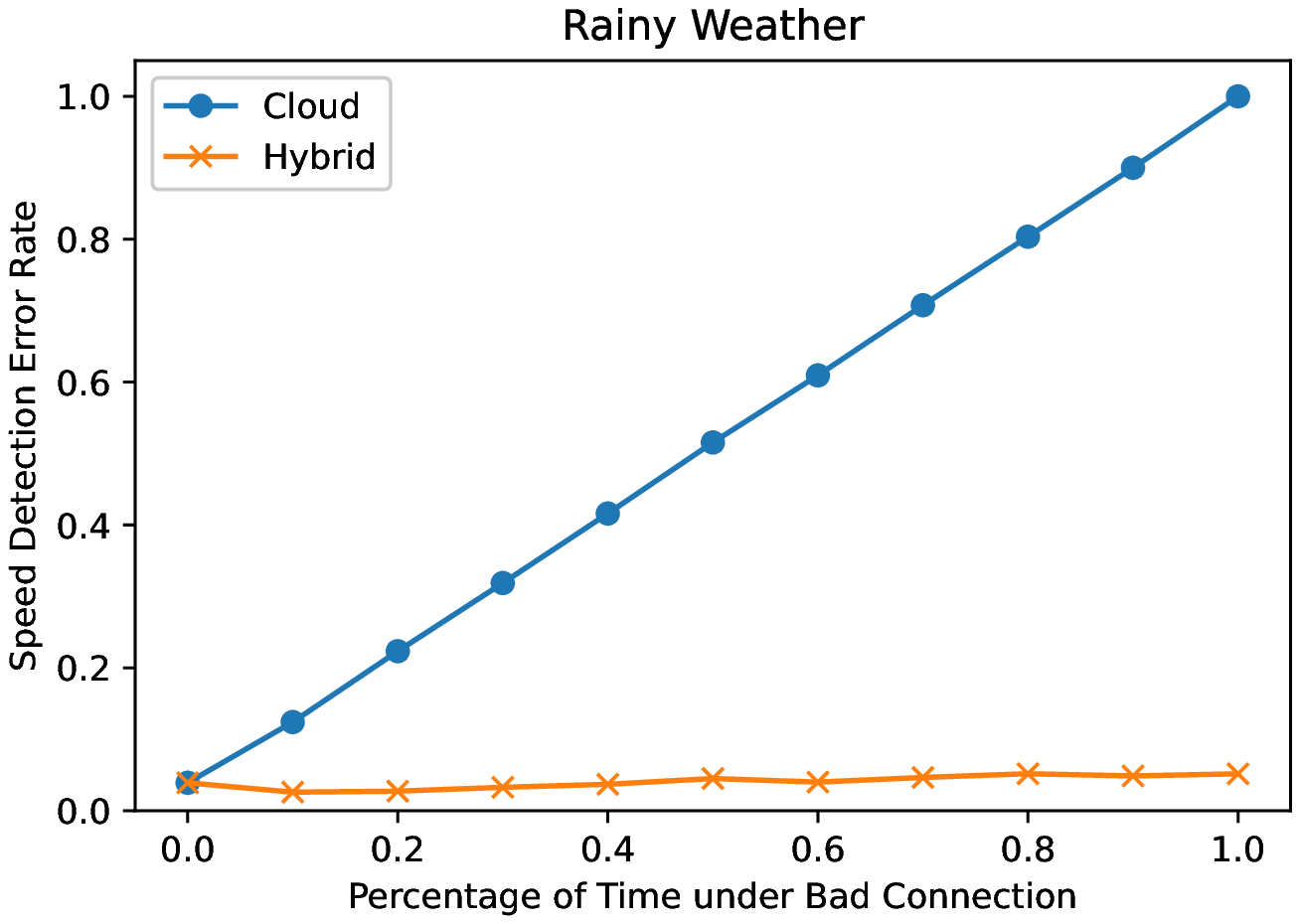}}
\end{minipage}
\begin{minipage}[c]{.32\textwidth}
    \centering
    \subfloat[$\epsilon_{S}$ in Snowy Day]{
        \includegraphics[width=\linewidth]{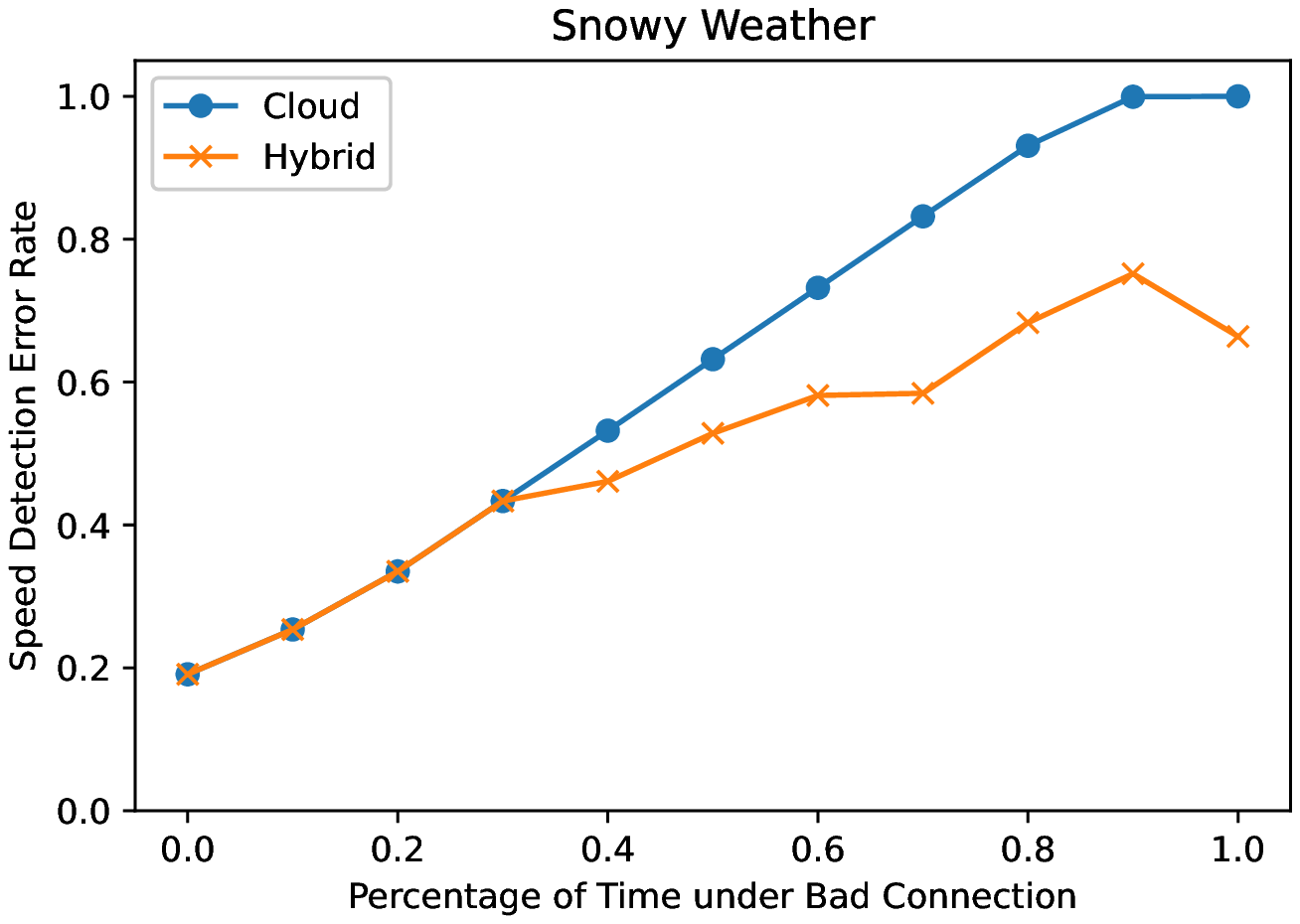}}
\end{minipage}
\begin{minipage}[c]{.32\textwidth}
    \centering
    \subfloat[$\epsilon_{RMS}$ in Sunny Day]{
        \includegraphics[width=\linewidth]{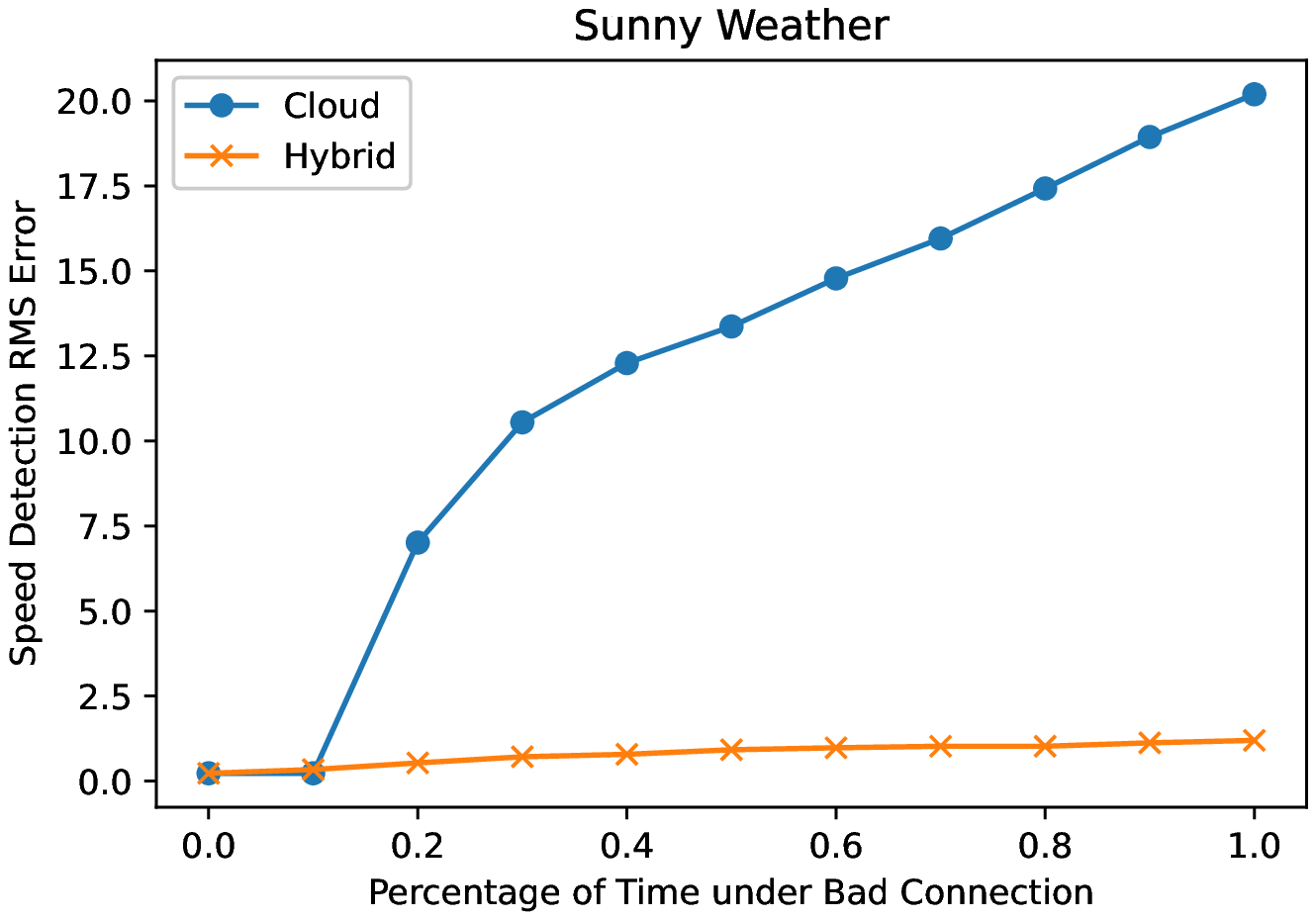}}
\end{minipage}
\begin{minipage}[c]{.32\textwidth}
    \centering
    \subfloat[$\epsilon_{RMS}$ in Rainy Day]{
        \includegraphics[width=\linewidth]{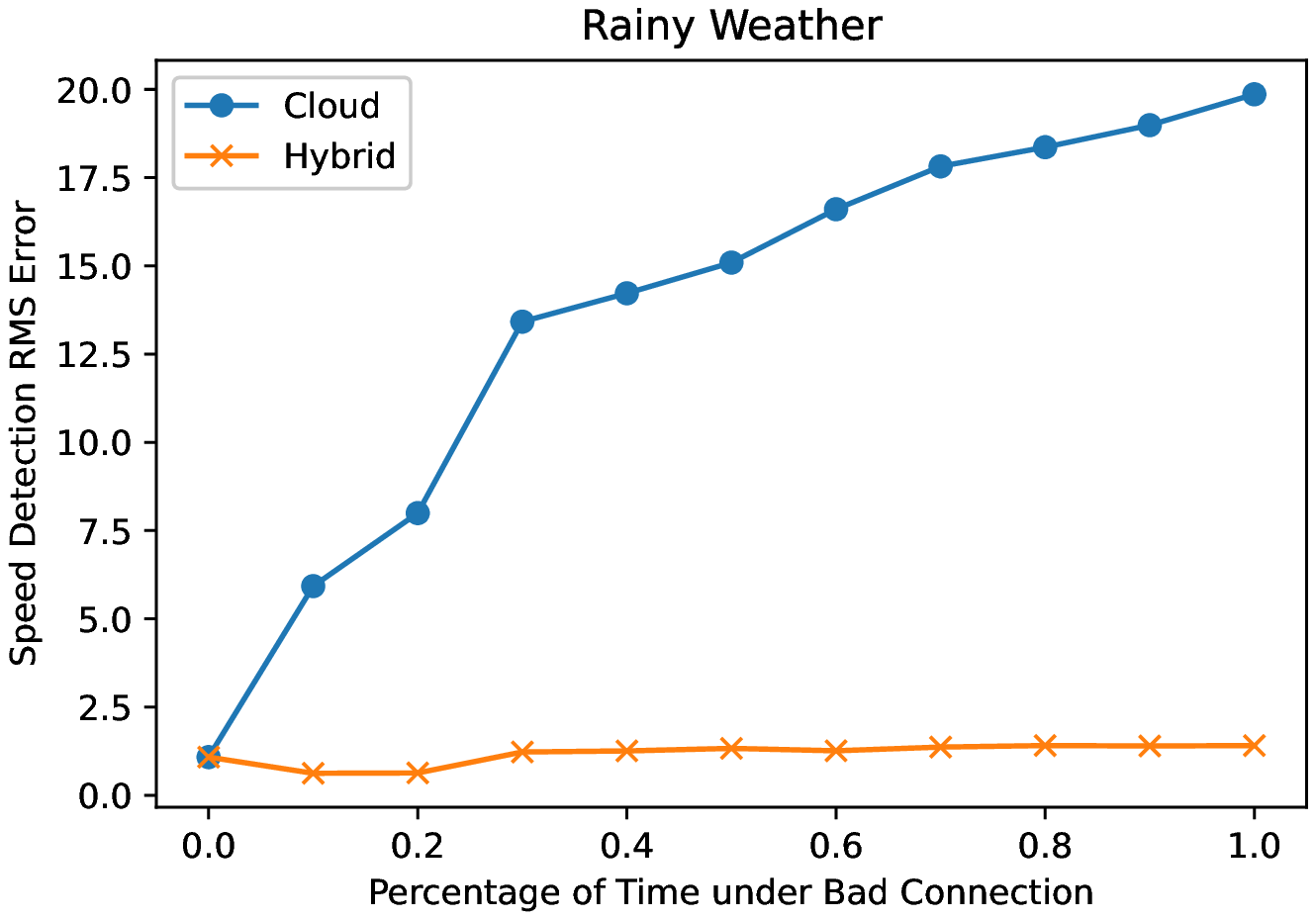}}
\end{minipage}
\begin{minipage}[c]{.32\textwidth}
    \centering
    \subfloat[$\epsilon_{RMS}$ in Snowy Day \label{fig:speed-inc}]{
        \includegraphics[width=\linewidth]{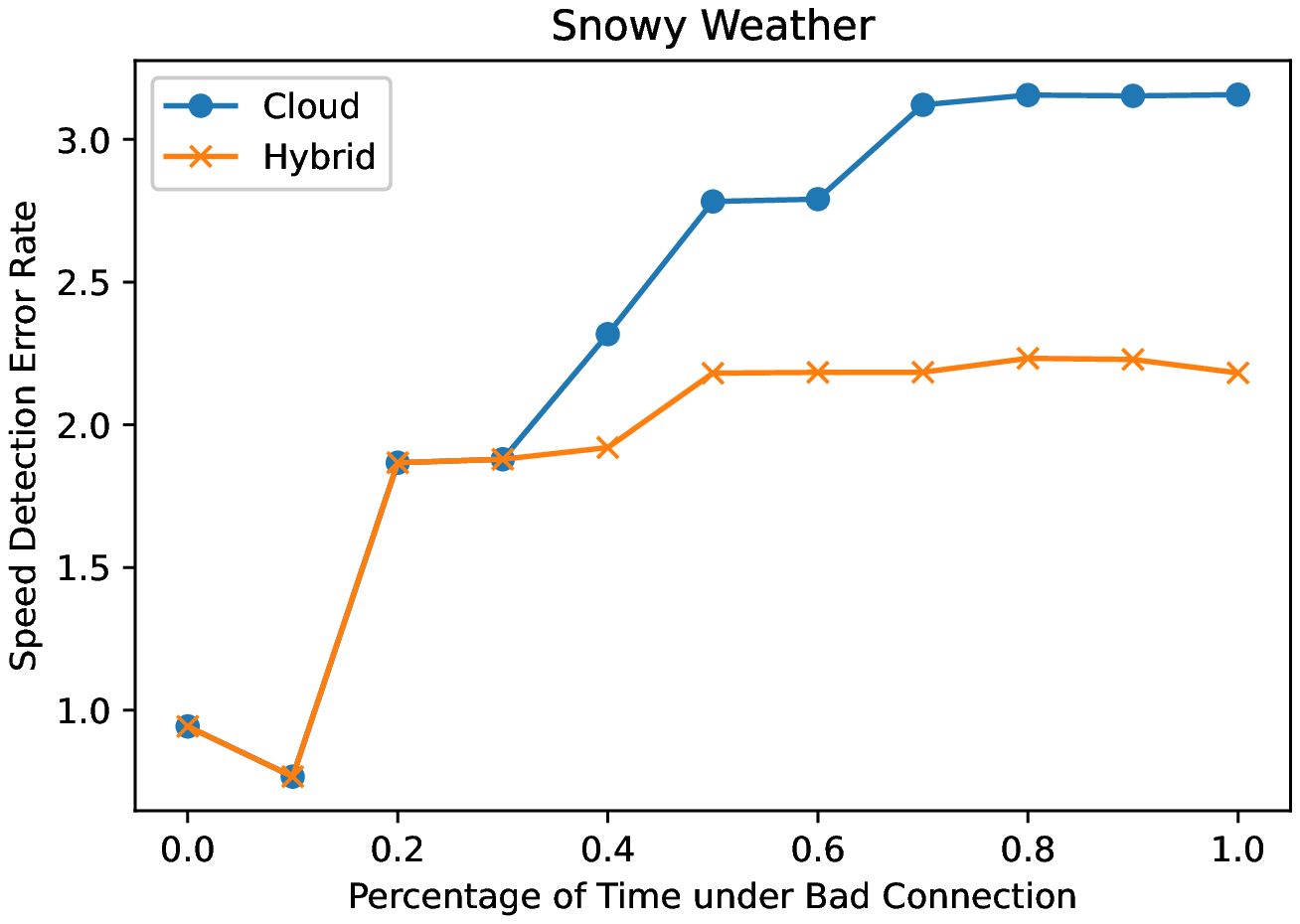}}
\end{minipage}
\vspace{-2mm}
\caption{$\epsilon_{S}$ (the 1$^{st}$ row) and $\epsilon_{RMS}$ (the 2$^{nd}$ row) under Different percentage of time for bad network condition}
\vspace{-5mm}
\label{fig:speed-slide}
\end{figure*}

\subsection{Comparison of the three strategies}


Based on our description in Section \ref{config}, we prepare two sets of configurations for each traffic monitoring task which consider the computing resources on the edge cloudlet and cloud servers. When the network condition is good, the video quality on the cloud side is good. The video processing algorithms can achieve relatively low values of $\epsilon_{C}$, $\epsilon_{S}$ and $\epsilon_{RMS}$. When the network condition is bad, the video on the cloud side is discontinuous with jumping frames. It is difficult for the video processing algorithms to detect the traffic conditions on the cloud side. Therefore, both the error rate and the RMS error of the cloud become very high. On the contrary, the video quality of the edge side is not affected by the network conditions, so that the video processing algorithms can maintain a relatively stable performance. It is worth mentioning that the overall evaluation results are summarized in Table \ref{tab:error-rate} with 95\% confidence interval. For $\epsilon_{C}$, we did not provide the confidence interval since it is a binary data. Also, since $\epsilon_{RMS}$ is the standard deviation of the residuals between measured speed and ground truth, the confidence interval is not provided.

Fig. \ref{fig:compare1} shows a comparison of the error rate for the Edge and Cloud schemes under different network and weather conditions. Under all weather conditions, the performance of both the traffic congestion detection and the vehicle speed detection with the Edge scheme is in between that of the Cloud scheme with good and bad network conditions. The bad weather conditions (i.e. snowy) increases the error rate of both the traffic congestion detection and the vehicle speed detection with the Edge and Cloud schemes because the rain and the snow blur the video and incur noise. When the network condition is bad, the cloud side is almost unable to detect the congestion and vehicle speed due to the jumping frames. As a result, we can see that the Cloud- bars are significantly higher than others in all situations.

In this work, our proposed two-tier computation model, Hybrid scheme, leverages the advantages of switching the computation between the Edge and the Cloud to achieve a more stable and accurate detection rate. The comparison of the Hybrid scheme with the Edge and Cloud schemes is presented in Fig. \ref{fig:compare2}. From this comparison, we observe three findings. (1) Under all weather conditions, the Hybrid scheme outperforms both the Edge and Cloud schemes since it performs dynamic switching between these two methods. (2) Under good weather condition, the performance of the Edge scheme is better than that of the Cloud scheme. The reason is that the jumping frame caused by network delay is the major issue. (3) Under bad weather condition (i.e. snowy), the performance of the Cloud scheme is better than that of Edge scheme. The reason is that the video quality is affected by the bad weather which makes the Edge scheme performing poorly in this scenario.

To understand the advantage of the proposed two-tier model, we design another set of evaluations to compare it with the Cloud scheme. During the evaluation, we calculate the average detection error rate within a sliding window. The sliding window changes from pure good network condition to pure bad network condition. The evaluation results are summarized in Fig. \ref{fig:congestion-slide} and Fig. \ref{fig:speed-slide}.

From the results, it is clear that detection error rates of the Cloud scheme on both tasks increase rapidly. 
Although the congestion detection error rate of the Cloud scheme decreases in the first two data points in Figure \ref{fig:speed-inc}, it is only a small fluctuation caused by the utilization of sliding window. From the high level point of view, the overall trend (i.e., detection error rates of the Cloud scheme increase rapidly with the ratios of bad network conditions) stays the same in all sub-figures.
Compared to the Cloud scheme, the detection error rates of the two-tier model increase at a relatively lower rate. Moreover, the increment stops or slows down after a certain percentage. Although both the Cloud scheme and the two-tier model show an increasing trend on detection error rate, the root causes are totally different. For the Cloud scheme, the increment in the error rate is caused by the increment in the percentage of time for the bad network condition. For the two-tier model, the increment in error rate is caused by the increasing reliance on results generated by the cloudlet. In Fig. \ref{fig:speed-slide}, we present both $\epsilon_{S}$ and $\epsilon_{RMS}$. These two sets of results can lead to the same comparison conclusion while the RMS error provides more detailed information. \guanxiong{Under all three different weather conditions, the error of the Hybrid scheme is at the same level as that of the Cloud scheme. However, when the percentage of time for bad network condition increases, the RMS error of Hybrid scheme becomes lower. The reason is that speed detection results from the Cloud are significantly affected by the bad network condition while the Hybrid scheme starts to rely on the cloudlet.}

\begin{table}[t]
    \caption{Encoding}
    \label{table:encoding}
    \begin{center}
    \begin{tabular}{ c | c | c | c }
    \hline \hline
    \multicolumn{2}{c |}{Weather} & \multicolumn{2}{c}{Network Condition} \\
    \hline
    Sunny \& Rainy & Snowy & Good & Bad \\
    \hline
    0 & 1 & 0 & 1 \\
    \hline \hline
    \end{tabular}
    \end{center}
\end{table}
\vspace{-3mm}
\subsection{Statistical Analysis}

 Based on the analysis of the experimental results done so far, we identify two major factors that affect the performance of the Edge and Cloud schemes. (1) The bad network condition significantly increases the error rate of the Cloud scheme on both congestion and speed detection. (2) The snowy weather makes the Edge scheme performing poorly on both congestion and speed detection. In order to provide a stronger support for these conclusions, we present statistical tests. We choose to report the analysis of variance (ANOVA) that provides a statistical test for the hypothesis testing \cite{scheffe1999analysis}. Through such hypothesis testing, we could determine if two variables are statistically correlated. As a result, we could utilize the ANOVA to statistically explore and validate the correlation between different error rates ($\epsilon_{C}, \epsilon_{S}, \text{and} ~ \epsilon_{RMS}$), network conditions, and weather conditions.

Since the weather information and network condition are categorical, we utilize the encoding method in Table \ref{table:encoding}. This encoding assigns both Sunny and Rainy weathers to the same default category since the ANOVA test results show that Rainy weather is not statistically significant to be separated from Sunny weather in terms of affecting both congestion and speed detection. The ANOVA results under this encoding setting are summarized in Table \ref{table:anova-connection} and \ref{table:anova-weather}.

\begin{table}[t]
    \caption{ANOVA (Error Rate \& Bad Network Condition)}
    \label{table:anova-connection}
    \begin{center}
    \begin{tabular}{ c | c | c | c }
    \hline \hline
    & & \multicolumn{2}{c}{Error Rate (Cloud)} \\
    \hline
    & & Congestion & Speed \\
    \hline
    \multirow{2}{*}{Bad Network Condition} & Coefficient & 0.6528 & 0.9496 \\
    & p\_value & $1.26e^{-12}$ & $4.43e^{-83}$ \\
    \hline \hline
    \end{tabular}
    \end{center}
\end{table}
\begin{table}[t]
    \caption{ANOVA (Error Rate \& Snowy Weather)}
    \label{table:anova-weather}
    \begin{center}
    \begin{tabular}{ c | c | c | c }
    \hline \hline
    & & \multicolumn{2}{c}{Error Rate (Edge)} \\
    \hline
    & & Congestion & Speed \\
    \hline
    \multirow{2}{*}{Snowy Weather} & Coefficient & 0.3125 & 0.4197 \\
    & p\_value & $1.53e^{-3}$ & $2.47e^{-23}$ \\
    \hline \hline
    \end{tabular}
    \end{center}
\end{table}

The coefficients are positive in Table \ref{table:anova-connection}, which means that bad network condition positively contributes to the error rate of the Cloud scheme on both congestion and speed detection. This means that changing from the good network condition to the bad network condition could enhance the error rate of the Cloud scheme on the two traffic monitoring tasks. Given that the p\_value of these two coefficients are much smaller than $0.01$, the positive relation between the error rate of the Cloud scheme and the bad network condition is statistically significant.

A similar conclusion can be drawn from the results in Table \ref{table:anova-weather}. The snowy weather positively contributes to the error rate of the Edge scheme on both congestion and speed detection. This means that changing from the sunny or rainy weather to the snowy weather could enhance the error rate of the Edge scheme on the two traffic monitoring tasks. Given the p\_value of these two coefficients are also smaller than $0.01$, the positive relation between the error rate of the Edge scheme and the snowy weather is also statistically significant.

To this point, we statistically show that the bad network condition affects the Cloud scheme severely while the snowy weather condition has a major effect on both the Edge and the Cloud schemes. We also indirectly show the advantage of our Hybrid scheme which can switch between the edge and the cloud based on the given network and weather conditions.
\vspace{-3mm}
\section{conclusion and future work}\label{sec:Conclude}

In this work, we design, implement, and analyze a small test-bed of the two-tier edge computing model for smart traffic monitoring system. This test-bed is deployed at the NJIT campus by recording real-world traffic on the road. To evaluate the proposed two-tier model, we conduct extensive measurements. Our evaluations include both congestion and speed detection objectives under 6 different combinations of network/weather conditions (good/bad network conditions and sunny/rainy/snowy weather condition). Based on the experiments, the proposed two-tier model outperforms both the individual Edge and Cloud schemes. Moreover, we identify that snowy weather and bad network condition are two major factors that increase the error rate of the Edge and Cloud schemes, respectively. Based on ANOVA tests, these relations are also shown to be statistically significant. Therefore, this analysis indirectly reflects the advantage of our two-tier model which can switch between the edge and the cloud based on the network and the weather conditions. In our future work, we will consider analyzing more traffic video analysis tasks, such as vehicle detection, vehicle frame detection, traffic accident detection, etc. We will also try to integrate some neural network based methods into our two-tier edge computing model to solve the traffic video analysis problems in the future.
\vspace{-3mm}
\bibliographystyle{IEEEtran}
\bibliography{ref}

\begin{IEEEbiography}[{\includegraphics[width=1in,height=1.25in,clip,keepaspectratio]{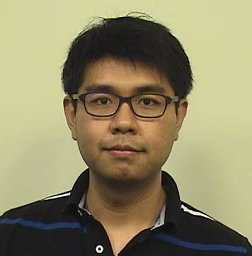}}]{Guanxiong Liu}
received the B.S. degree in electrical and computer engineering from the Southeast University in 2013, the M.S. degree in electrical and computer engineering from Worcester Polytechnic Institute in 2015. Then, he joined Intel as a full time Verification Engineer for over a year. He is currently a Ph.D. student with Prof. Abdallah Khreishah in the Department of Electrical and Computer Engineering, New Jersey Institute of Technology. His current research interests include machine learning security, big data analysis, and smart city. He is also a Data Scientist Summer Intern in Cylera, Inc.
\end{IEEEbiography}
\begin{IEEEbiography}[{\includegraphics[width=1in,height=1.25in,clip,keepaspectratio]{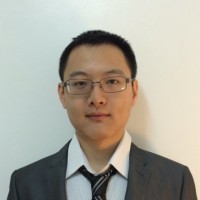}}]{Hang Shi}
received the B.S. degree in automation from University of Science and Technology of China, Hefei, Anhui, China in 2012, and the M.S. degree in electrical engineering from New York University, Brooklyn, NY, USA in 2014. He is currently pursuing the Ph.D. degree with the Department of Computer Science, New Jersey Institute of Technology, Newark, NJ, USA. His current research interests include computer vision, image and video analysis, machine learning, pattern recognition, and image classification.
\end{IEEEbiography}
\begin{IEEEbiography}[{\includegraphics[width=1in,height=1.25in,clip,keepaspectratio]{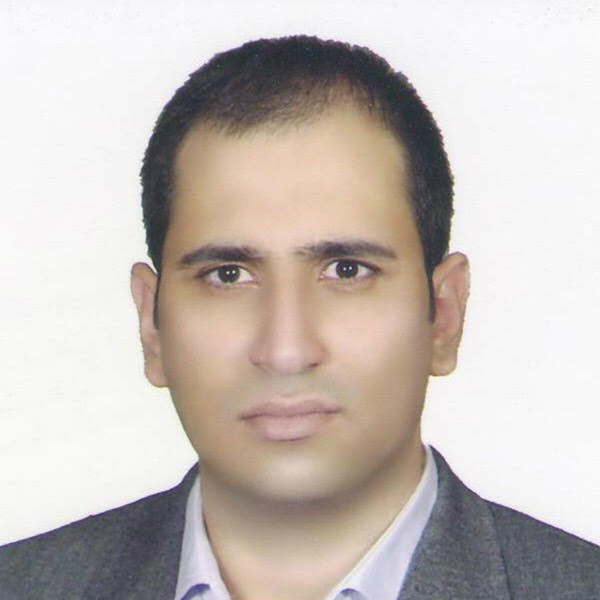}}]{Abbas Kiani}
(S'14) received the B.Sc. degree in electrical engineering from Imam Khomeini International University, Qazvin, Iran, the M.Sc. degree in communication engineering from Shahed University, Tehran, Iran, and the Ph.D. degree in electrical engineering from New Jersey Institute of Technology, Newark, NJ, USA. His current research interests include 5G, edge computing, and network optimization.
\end{IEEEbiography}
\begin{IEEEbiography}[{\includegraphics[width=1in,height=1.25in,clip,keepaspectratio]{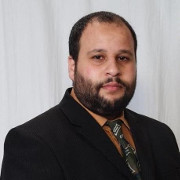}}]{Abdallah Khreishah}
received his Ph.D and M.S. degrees in Electrical and Computer Engineering from Purdue University in 2010 and 2006, respectively. Prior to that, he received his B.S. degree with honors from Jordan University of Science \& Technology in 2004. During the last year of his Ph.D, he worked with NEESCOM. In Fall 2012, he joined the Electrical and Computer Engineering department of NJIT as an Assistant Professor and promoted to Associate Professor in 2017. His research spans the areas of wireless networks, visible-light communication, vehicular networks, congestion control, cloud \& edge computing, and network security. His research projects are funded by the National Science Foundation, New Jersey Department of Transportation, and the UAE Research Foundation. He is currently serving as an associate editor for several International Journals. He served as the TPC chair for WASA 2017, IEEE SNAMS 2014, IEEE SDS-2014, BDSN-2015, BSDN 2015, IOTSMS-2015. He has also served on the TPC committee of several international conferences such as IEEE Infocom. He is a senior member of IEEE and the chair of the IEEE EMBS North Jersey chapter.
\end{IEEEbiography}
\begin{IEEEbiography}[{\includegraphics[width=1in,height=1.25in,clip,keepaspectratio]{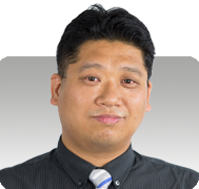}}]{Joyoung Lee}
received the B.S. degree in transportation engineering from Hanyang University, Ansan, South Korea, in 2000, and the M.S. and Ph.D. degrees in civil and environmental engineering from the University of Virginia at Charlottesville, in 2007 and 2010, respectively. He was the Laboratory Manager at the Saxton Transportation Operations Laboratory, Federal Highway Administration Turner-Fairbank Highway Research Center. He is currently an Associate Professor with the John A. Reif, Jr. Department of Civil and Environmental Engineering, New Jersey Institute of Technology. Dr. Lee’s research interest lies in the developments and evaluations of diverse Intelligent Transportation Systems (ITS) applications covering Connected Automated Vehicle(CAV)-based intersection controls, advanced traveler information systems, smart traffic congestion sensing, and advanced ITS modeling. He is a member of the Transportation Research Board Travel time, Speed, and Reliability Subcommittee. He is an Associate Editor of the KSCE Journal of Civil Engineering.
\end{IEEEbiography}
\newpage
\begin{IEEEbiography}[{\includegraphics[width=1in,height=1.25in,clip,keepaspectratio]{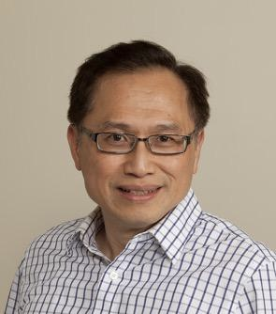}}]{Nirwan Ansari}
(S'78-M'83-SM'94-F'09), Distinguished Professor of Electrical and Computer Engineering at the New Jersey Institute of Technology (NJIT), received a Ph.D. from Purdue University, an MSEE from the University of Michigan, and a BSEE (summa cum laude with a perfect GPA) from NJIT.

He has (co-)authored three books and more than 600 technical publications, over 320 published in widely cited journals/magazines. He has also been granted more than 40 U.S. patents. His current research focuses on green communications and networking, cloud computing, drone-assisted networking, and various aspects of broadband networks. Some of his recognitions include several Excellence in Teaching Awards, a few best paper awards, the NCE Excellence in Research Award, several ComSoc TC technical recognition awards, the NJ Inventors Hall of Fame Inventor of the Year Award, the Thomas Alva Edison Patent Award, Purdue University Outstanding Electrical and Computer Engineering Award, the NCE 100 Medal, and designation as a COMSOC Distinguished Lecturer.
\end{IEEEbiography}
\begin{IEEEbiography}[{\includegraphics[width=1in,height=1.25in,clip,keepaspectratio]{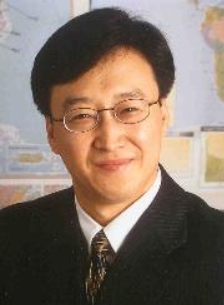}}]{Chengjun Liu}
is a Professor of Computer Science and the Director of the Face Recognition and Video Processing Laboratory at the New Jersey Institute of Technology, Newark, NJ, USA. He has developed the evolutionary pursuit method, the enhanced Fisher models, the Gabor Fisher classifier, the Bayesian discriminating features method, the kernel Fisher analysis method, new color models, new image descriptors, and new similarity measures. His current research interests include pattern recognition, machine learning, computer vision, image and video analysis, and security.
\end{IEEEbiography}
\begin{IEEEbiography}[{\includegraphics[width=1in,height=1.25in,clip,keepaspectratio]{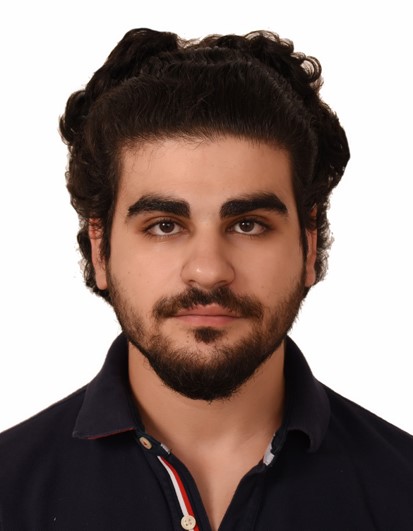}}]{Mustafa Mohammad Yousef}
has a master’s degree in electrical engineering from New Jersey Institute of Technology and a bachelor’s degree in mechanical engineering from American University of Sharjah. During his master’s degree he worked on semiconductors technologies with a focus on thin film transistors and then he started investigating the field of machine learning and data science.
\end{IEEEbiography}

\end{document}